\documentclass[3p,times,twocolumn]{elsarticle}
\usepackage{amsmath,amssymb,amsfonts}
\usepackage{graphicx}
\usepackage{textcomp}
\usepackage{xcolor}
\usepackage{url}
\usepackage{bm}
\usepackage{bbm}
\usepackage{algorithm}
\usepackage{algorithmic} 
\usepackage{multirow}
\usepackage{caption}
\usepackage{subcaption}
\usepackage{supertabular}
\usepackage{lipsum}
\usepackage{graphicx}
\usepackage{makecell}
\usepackage{float}

\begin{document}

\begin{frontmatter}

\title{Comparative Evaluation of Recent Universal Adversarial Perturbations \\ in Image Classification}

\author[1]{Juanjuan Weng}
\ead{wengjuan@stu.xmu.edu.cn}
\author[1]{Zhiming Luo\corref{cor1}}
\ead{zhiming.luo@xmu.edu.cn}
\author[1]{Dazhen Lin}
\ead{dzlin@xmu.edu.cn}
\author[1]{Shaozi Li}
\ead{szlig@xmu.edu.cn}
\address[1]{Department of Artificial Intelligence, Xiamen University, Xiamen 361005, China}
\cortext[cor1]{Corresponding author}

\begin{abstract}
The vulnerability of Convolutional Neural Networks (CNNs) to adversarial samples has recently garnered significant attention in the machine learning community. Furthermore, recent studies have unveiled the existence of universal adversarial perturbations (UAPs) that are image-agnostic and highly transferable across different CNN models. In this survey, our primary focus revolves around the recent advancements in UAPs specifically within the image classification task.
We categorize UAPs into two distinct categories, \textit{i.e.}, noise-based attacks and generator-based attacks, thereby providing a comprehensive overview of representative methods within each category. By presenting the computational details of these methods, we summarize various loss functions employed for learning UAPs.
Furthermore, we conduct a comprehensive evaluation of different loss functions within consistent training frameworks, including noise-based and generator-based. The evaluation covers a wide range of attack settings, including black-box and white-box attacks, targeted and untargeted attacks, as well as the examination of defense mechanisms.
 Our quantitative evaluation results yield several important findings pertaining to the effectiveness of different loss functions, the selection of surrogate CNN models, the impact of training data and data size, and the training frameworks involved in crafting universal attackers. Finally, to further promote future research on universal adversarial attacks, we provide some visualizations of the perturbations and discuss the potential research directions.
\end{abstract}

\end{frontmatter}

\section{Introduction}
In the past few years, Deep Convolutional Neural Networks (CNNs) have achieved remarkable performance in various computer vision tasks, such as image classification~\cite{szegedy2015going,huang2017densely}, object detection~\cite{ren2015faster,redmon2018yolov3} and image segmentation~\cite{chen2017deeplab,long2015fully}. Despite the achievements, recent studies~\cite{szegedy2013intriguing,goodfellow2014explaining}  have revealed a critical vulnerability of CNNs to adversarial examples. These adversarial examples refer to the addition of quasi-imperceptible perturbations to an image, leading to wrongly predictions of CNNs. The existence of adversarial perturbations poses a significant threat to the real-world applications of CNNs, particularly in domains such as speech recognition~\cite{wang2020towards,wu2020audio}, facial verification systems~\cite{zhong2020towards,dong2019efficient} and image classification~\cite{wang2022transferable,peng2021ensemblefool}. 

Recently, many per-instance adversarial attacks~\cite{xie2019improving,dong2019evading,liu2016delving} have been proposed to generate disruptive adversarial examples by training with the FGSM-related techniques~\cite{goodfellow2014explaining,kurakin2016adversarial,dong2018boosting}. In FGSM-based methods, perturbations are individually optimized for each input image to produce its corresponding adversarial sample. These adversarial samples exhibit strong transferability, enabling them to effectively attack diverse CNN models with high success rates. In contrast to per-instance perturbations, Moosavi-Dezfooli and Fawzi~\cite{moosavi2017universal} further found the existence of image-agnostic adversarial perturbations, in which a single universal adversarial perturbation (UAP) can fool the CNN model on a majority of input samples. The UAPs are more efficient in terms of computation cost when compared with the per-instance adversarial attacks. 

% A short description of the development of the UAP
Following the findings of~\cite{moosavi2017universal}, several different methods have been proposed for crafting UAPs from different aspects. Mopuri et al.~\cite{mopuri2017fast} train the UAP without using any training data by wrongly activating the neurons in the CNNs. Besides, there are some attempts to cross the correct decision boundaries by using different training loss functions, such as, cross-entropy-based~\cite{poursaeed2018generative,naseer2019cross}, margin-based~\cite{zhang2020understanding}, similarity-based~\cite{zhang2021data}. On the other hand, the Generative Adversarial Networks (GANs)~\cite{poursaeed2018generative} also have been introduced for learning the UAPs~\cite{poursaeed2018generative,hayes2018learning,naseer2021generating}. Instead of just tricking the CNNs into making any incorrect predictions, previous studies~\cite{zhang2020understanding,naseer2021generating} found that the universal attacks can misdirect the CNNs to wrongly classify the adversarial samples into a pre-defined class, also known as the targeted attack. Similar to the per-instance adversarial perturbation, the UAPs also show strong transferability in attacking other unknown black-box CNN models.

\begin{figure}[t]
    \centering
    \includegraphics[width=1\linewidth]{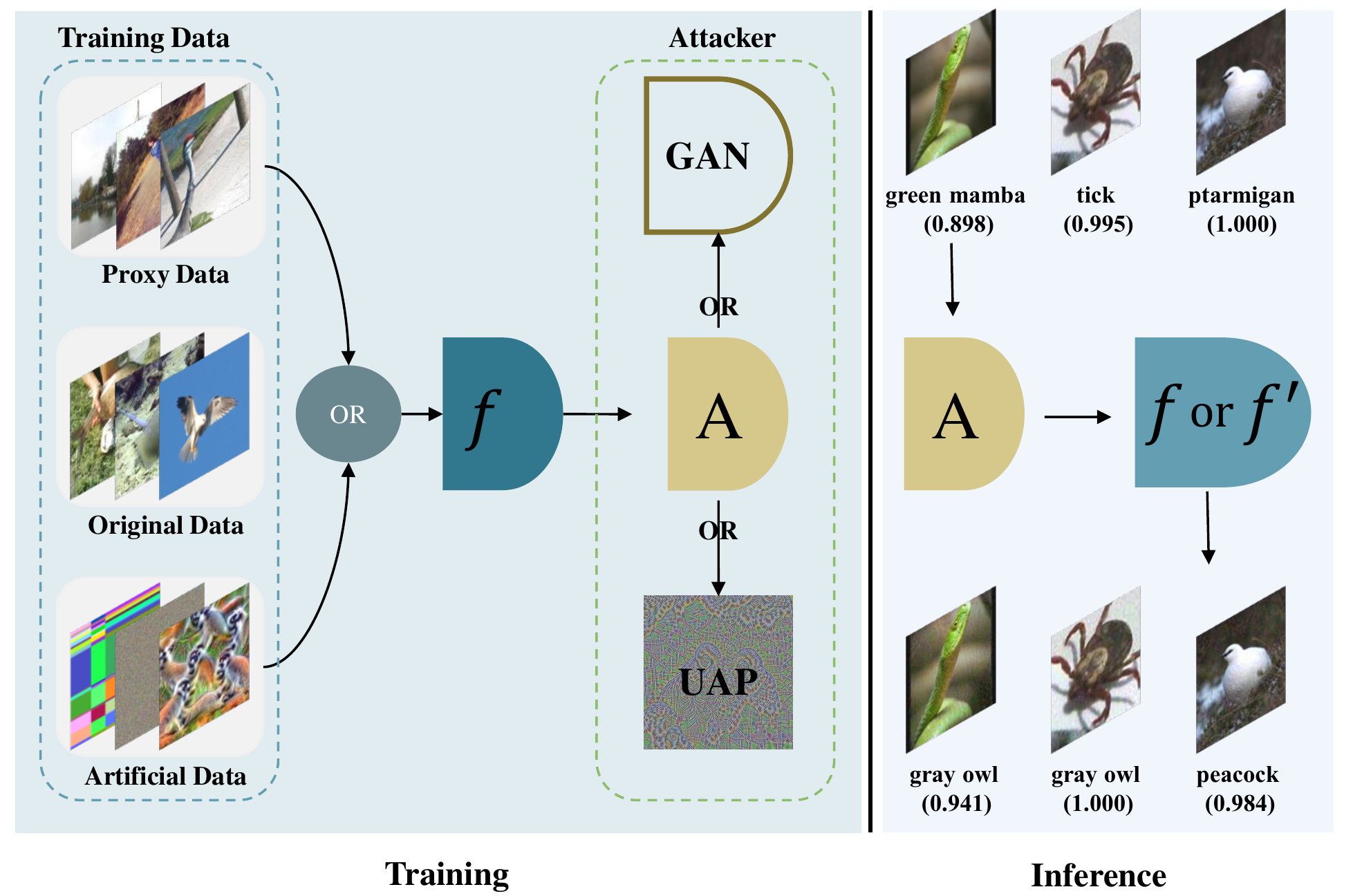}
    \caption{Procedure of universal attack methods in deep image classifiers. During training, the universal attacker (generator-based or universal adversarial perturbations) is learned from training data (data-driven,data-independent) and CNN model $f$. During inference, the attacker (A) is added on almost all the input images $X$ to craft adversarial examples fooling the white-box model $f$ or unknown target model $f'$.}
    \label{fig:defined}
\end{figure}

\begin{table*}[t]
\centering
\caption{Attribute of different universal attacking methods.}
\label{tab:survey_list}
\resizebox{1\linewidth}{!}{
\begin{tabular}{l|l|c|c|c|c}
\hline\hline
\textbf{Type}  &\textbf{Method} & Venue & \textbf{Source Data} & \textbf{Attack Type} & \textbf{Perturbation Norm} \\ \hline\hline
\multirow{10}{*}{Noise-based} 
& UAP~\cite{moosavi2017universal} & CVPR-2017 & $\checkmark$ & non-targeted   & $l_2,l_{\infty}$ \\ 
 & SPM~\cite{khrulkov2018art} & CVPR-2018 & $\checkmark$ &  non-targeted  & $l_{\infty}$ \\ 
 & DTA~\cite{benz2020double} &ACCV-2020 & $\checkmark$ &  targeted  & $l_{\infty}$ \\ 
 & CD-UAP~\cite{zhang2020cd} &AAAI-2020 & $\checkmark$   &targeted  & $l_{\infty}$  \\ 
% \cline{2-6}
& FFF~\cite{mopuri2017fast} &BMVC-2017 & - & non-targeted & $l_{\infty}$  \\
& GD-UAP~\cite{mopuri2018generalizable} & TPAMI-2018& - & non-targeted& $l_{\infty}$\\
&  PD-UAP~\cite{liu2019universal} &ICCV-2019 & - & non-targeted& $l_{\infty}$\\
& F-UAP~\cite{zhang2020understanding} &CVPR-2020 & - &non-targeted/targeted &$l_{\infty}$ \\
& Jigsaw-UAP~\cite{zhang2021data} &ICCV-2021 & - & non-targeted/targeted& $l_{\infty}$ \\ 
\hline\hline
\multirow{10}{*}{Generator-based } 
& GAP~\cite{poursaeed2018generative}& CVPR-2018 & $\checkmark$ &non-targeted/targeted&$l_2,l_{\infty}$\\
&  GM-UAP~\cite{hayes2018learning} & SPW-2018 & $\checkmark$ &non-targeted/targeted &$l_2,l_{\infty}$ \\
&  NAG~\cite{mopuri2018nag} &CVPR-2018  & $\checkmark$ &non-targeted &$l_{\infty}$ \\
&  GM-TUAP~\cite{hashemi2020transferable} & Arxiv-2020 & $\checkmark$&non-targeted/targeted&$l_2,l_{\infty}$\\
&  TTP~\cite{naseer2021generating} &ICCV2021 & $\checkmark$ &targeted & $l_{\infty}$ \\
&  GAP++~\cite{mao2020gap++} &IJCAI-2019  & $\checkmark$ &targeted &$l_1,l_2,l_{\infty}$ \\ 
% \cline{2-6}
& AAA~\cite{mopuri2018ask} & ECCV-2018 &- & non-targeted& $l_1$ \\ 
&  CD-TAP~\cite{naseer2019cross} &NeurIPS-2019 &- &non-targeted/targeted & $l_{\infty}$\\
\hline\hline
\end{tabular}
}
\end{table*}

% How we group this survey
To have an overview of the large body of recently published literature, in this paper, we review the development of UAPs in the context of image classification. As presented in Table~\ref{tab:survey_list}, we have categorized the methods into two main groups: noise-based methods~\cite{moosavi2017universal,khrulkov2018art,benz2020double,zhang2020cd} and generator-based methods~\cite{poursaeed2018generative,hayes2018learning,naseer2021generating}. 
% The noise-based methods optimize the UAP directly, while the generator-based methods train a separate generative network to obtain the perturbation. 
The key distinction between noise-based and generator-based methods and methods is that the former directly optimizes the UAP by updating the perturbation, while the latter relies on training a UAP generation network to obtain the perturbation indirectly. %The drawback of generator-based methods is the requirement of a generative model, while noise-based methods offer a more direct and efficient optimization approach.
% Additionally, regarding the utilization of source data, these methods  can also be further categorized into data-driven and data-independent methods. 
The overall computation procedures of these methods are depicted in Fig.~\ref{fig:defined}. As shown, noise-based methods train a universal adversarial perturbation, while generator-based methods employ a generative adversarial network (GAN) to generate the perturbations. Furthermore, the availability of the original training data is often limited in practical cases, several approaches utilize proxy data, artificial data, or even no data to learn the universal attackers for addressing this issue. 
% Moreover, several methods can be employed to learn both untargeted and targeted adversarial examples by utilizing different training loss functions.
% data-driven methods rely on the original source training data, whereas data-independent methods leverage proxy or artificial data, or even no data, for the training of the universal attackers.

% Experiments
Despite the rapid progress of the UAPs, we observe that various loss functions used for learning the UAP are adopted in different training frameworks. However, there is no systematic comparison of the loss functions under the same training settings. To gain a deeper understanding of the characteristics of different loss functions in UAPs, we conduct comprehensive large-scale experiments to evaluate their attacking performance. These experiments are conducted under both the noise-based and generator-based frameworks. Furthermore, considering the unavailability of the source data, we utilize proxy data, such as Microsoft COCO~\cite{lin2014microsoft}, to craft the UAPs. We also compare the performance of UAPs generated using proxy data with those generated using the original source data under the same training frameworks. Finally, we utilize the UAPs trained with different loss functions to assess their performance in various attack scenarios, including black-box and white-box attacks, as well as targeted and untargeted attacks. Furthermore, we examine the effectiveness of these UAPs against defense mechanisms.

By analyzing the quantitative results, we have some important observations of the UAPs. \textbf{1)} There is no significant difference in the fooling rate of the noise-based framework and the generator-based framework. \textbf{2)} The training data barely influences the
attacking performance. Either training on the source dataset
ImageNet or Microsoft COCO (proxy dataset) results in similar performance under the same loss function. \textbf{3)} The simplest similarity-based $L_{cos}$, which minimizes the cosine distance of logits between the original sample and its adversarial counterparts, can achieve the overall highest fooling rate under different training frameworks in non-targeted attacks. 
The margin-based loss consistently achieves a higher success rate for both targeted and non-targeted attacks. Besides, different cross-entropy-based loss functions used in the targeted attack have similar performance. 
\textbf{4)} Compared with the VGG~\cite{simonyan2014very}, ResNet~\cite{he2016deep}, DenseNet~\cite{huang2017densely} based network architecture, attacking the GoogleNet~\cite{szegedy2015going} with inception modules is with the relatively lower transfer fooling rate for both non-targeted and targeted attacks.
\textbf{5)} The adversarially trained CNNs based on the $l_\infty$-trained adversarial samples show better robustness for defending against various types of UAPs. 

Different from the existing surveys in the literature on UAP~\cite{zhang2021survey,chaubey2020universal} that primarily summarize the recent progress on universal adversarial attacks across different domains such as image, audio, video, and text, our survey focuses specifically on UAPs in the field of image classification. We categorize the existing UAPs into noise-based and generator-based methods and examine the different loss functions used for learning UAPs. Especially, we conduct comprehensive experiments to systematically analyze the performance of these loss functions under consistent training settings across various attack scenarios. Besides, we provide several important findings pertaining to the effectiveness of different loss functions, the selection of surrogate CNN models, the impact of training data and data size, and the training frameworks involved in crafting universal attackers. Finally, we provide
visualizations of the perturbations and discuss potential research directions to further promote future research on universal adversarial attacks. 

The structure of this survey is organized as follows. In Section~\ref{sec:def}, we describe the basic mathematics definition of UAP and the corresponding evaluation metrics for evaluating the attacking performance. In Section~\ref{sec:method}, we review the computation details of different UAP methods. In Section~\ref{sec:experiment}, we comprehensively evaluate the performance of using different loss functions for crafting the UAP under noise-based and generator-based frameworks. We provide some visualization and discuss the future directions in Section~\ref{sec:future}. Finally, we conclude this paper in Section~\ref{sec:conclusion}.

\section{Definition and Evaluation Metrics}
\label{sec:def}
\subsection{Definition of UAPs in deep image classifiers}
% The existence of universal adversarial perturbations is important, which can be used to fool a target model $f$ on all the images in the target data distribution $X \in \mathbf{R^d}$. And the UAPs as first demonstrated in~\cite{moosavi2017universal} can cause misclassification of the adversarial example $(x+\delta)$ to fool a target model $f$ when added to a sample image $x\in X$.

Given a CNN model $f$, the UAP~\cite{moosavi2017universal,khrulkov2018art} is an image-agnostic perturbation $\delta$ which will cause misclassification of the adversarial example $(x+\delta)$ when added to an image $x$. The main objective is to find such perturbation $\delta$ that can fool the $f$ on almost all the input images belonging to the target data distribution $X$. Additionally, $\delta$ should be sufficiently small to be imperceptible, which commonly is constrained by a pre-defined upper-bound $\epsilon$ on the $l_p$-norm, commonly termed as $\left \| \cdot \right \|_p$. Formally, in the non-targeted attack, we can summarize the main objective of the UAPs as:
\begin{equation}
\label{uap_def}
\begin{aligned}
  \mathbbm{P}\big(f(x+\delta)\big) & \neq \mathbbm{P}\big(f(x)\big), \mbox{ for most $x\sim X$}, \\
  \mbox{s.t.} \quad ||\delta||_p & \leq \epsilon,
\end{aligned}
\end{equation}
where $f(x)$ and $f(x+\delta)$ denote the output probabilities after the softmax related to the input image $x$ and the adversarial example $x+\delta$ computed by model $f$, $\mathbbm{P}\big(\cdot\big)$ computes the corresponding predicted label $c=\arg\max f(x)$.

In the targeted attack, we aim to learn the $\delta$ that fool the $f$ classifying $x+\delta$ to a specific targeted class $t$, instead of any incorrect class. The corresponding object function is denoted as follows,
\begin{equation}
\label{tuap_def}
\begin{aligned}
  \mathbbm{P}\big(f(x+\delta)\big) & = t, \mbox{ for most $x\sim X $}, \\
  \mbox{s.t.} \quad ||\delta||_p & \leq \epsilon, 
\end{aligned}
\end{equation}
where $t$ is a specifically targeted class. In this study, we conduct the evaluation by setting $p=\infty$ and $\epsilon= 10$ for images in the range of $[0,255]$ as in~\cite{moosavi2017universal,mopuri2018generalizable,poursaeed2018generative}.

\subsection{Metrics for evaluating the UAPs}
Given the above definition of UAPs, we introduce the widely used metrics for evaluating the effectiveness of the crafted UAP in non-targeted and targeted attacks, respectively. 

\subsubsection{Non-targeted attacks} 
In the case of non-targeted UAPs, the non-targeted fooling rate (ntFR)~\cite{moosavi2017universal,mopuri2018generalizable} is a common metric for evaluation. Given a target dataset $\mathcal{X}$ and a target model $f$, the ntFR is the percentage of images whose prediction changes when the UAP is added,
\begin{equation}
    \mbox{ntFR} = \sum_{x \in \mathcal{X}} \frac{\mathbbm{1} \Big(\mathbbm{P}\big(f(x+\delta)\big) \neq \mathbbm{P}\big(f(x)\big)\Big)}{|\mathcal{X}|},
\end{equation}
where $|\mathcal{X}|$ is the number of images in the testing dataset, $\mathbbm{1}$ represents the indicator function, which takes the value of 1 when the condition $\mathbbm{P}\big(f(x+\delta)\big) \neq \mathbbm{P}\big(f(x)\big)$ is true, and 0 otherwise. Besides, the transfer fooling rate is adopted to evaluate the transferability of $\delta$ learned from $f$ by attacking an unknown black model $f'$, which is computed by 
$\mbox{ntFR} = \sum_{x \in \mathcal{X}} \frac{\mathbbm{1} \Big(\mathbbm{P}\big(f'(x+\delta)\big) \neq \mathbbm{P}\big(f'(x)\big)\Big)} {|\mathcal{X}|}$.

\subsubsection{Targeted attacks}

In the case of targeted UAPs, the targeted fooling rate (tFR) metric is used for evaluation by computing the percentage of adversarial samples (except the samples of targeted class) to a pre-defined target class $t$~\cite{mopuri2018generalizable,zhang2020understanding}. The computation of tFR is 
\begin{equation}
    \mbox{tFR}  = \sum_{x \in (\mathcal{X}-\mathcal{X}_t)} \frac{\mathbbm{1} \Big(\mathbbm{P}\big(f(x+\delta)\big) = t\Big)}{|\mathcal{X}|-|\mathcal{X}_t|},
\end{equation}
where $t$ is the pre-defined targeted class, $(\mathcal{X}-\mathcal{X}_t)$ represents all test samples excluding the target samples,
and $|\mathcal{X}_t|$ is the number of targeted class samples. Additionally, the ntFR is also used for evaluation in the targeted UAPs. The transfer targeted fooling rate for attacking the black-box model $f'$ is used for evaluating the targeted transferability.

\section{Universal Adversarial Attack Methods}
\label{sec:method}
In this section, we will provide a detailed explanation of the computational procedures involved in both noise-based attack methods and generator-based attack methods used for crafting the universal adversarial perturbations.

\subsection{Noise-based attack methods}
The noise-based attack methods directly train a universal adversarial perturbation that can be applied to all input images. These methods aim to deceive the target model with a high success rate. Representative methods using the source data include UAP~\cite{moosavi2017universal}, SPM~\cite{khrulkov2018art}, DT-UAP~\cite{benz2020double}, and CD-UAP~\cite{zhang2020cd}. However, in practical cases, the availability of the original training data is often limited. To address this issue, there have been attempts~\cite{mopuri2017fast,mopuri2018generalizable,zhang2021data} to employ the proxy data, artificial data, or even no data for crafting the UAP.

\subsubsection{UAP}
% Universal Adversarial Perturbations
Moosavi-Dezfooli et al.~\cite{moosavi2017universal} propose the first UAP method for fooling the CNN models. To compute the UAP, Moosavi-Dezfooli et al.~\cite{moosavi2017universal} introduce an iterative algorithm to find the UAP, and the overall optimization procedure is illustrated in Algorithm~\ref{alg:UAP}. Given a set of training images $X$, the algorithm iteratively computes the perturbation $\Delta \delta_{i}$ of each sample to make an adversarial example cross the decision boundary of the real predicted category. Then, the $\Delta \delta_{i}$ will be aggregated to update the final $\delta$. During the optimization, the $\delta$ will be projected to obey constraints within the $\epsilon$ of $l_p$. Finally, the iteration will be terminated until the ``fooling rate" exceeds the pre-defined threshold $\eta$.

\begin{algorithm}[t]
  \caption{Computation of universal perturbations. }
  \label{alg:UAP}
  \textbf{Inputs:} Training data $X$, a CNN model $f$, the perturbation budget $\epsilon$ under the $l_p$ norm, the desired fooling rating $\eta$.\\
  \textbf{Outputs:} Universal perturbation $\delta$.
  
  \begin{algorithmic}[1]
    \STATE Initialize $\delta\leftarrow 0$.
        \WHILE{$\sum{\big(\mathbbm{P}\big(f(x+\delta)\big) \neq \mathbbm{P}\big(f(x)\big)\big)} \leq \eta$}
            \FOR{each sample $x_i\in X$}
               \IF{$\mathbbm{P}\big(f(x+\delta)\big) = \mathbbm{P}\big(f(x)\big)$}
               \STATE $c = \arg\max f(x_i)$
               \FOR{$k \neq c$}
               \STATE $w_k' \leftarrow \nabla f_k(x_i+\delta ) - \nabla f_{c}(x_i+\delta) $
               \STATE $f_k' \leftarrow  f_k(x_i+\delta ) -  f_{c}(x_i+\delta) $
        %       \STATE Compute the minimal perturbation that sends $x_i+\delta$ to the decision boundary:\\
        % $\Delta v_i \leftarrow \arg\min_{r} ||\delta||_2\ s.t.\ f(x_i+\delta+r)\neq f(x_i)$.
               \\
              \ENDFOR
            \STATE  $\hat{l} \leftarrow \arg\min_{k\neq c} \frac{\left|f_{k}^{'}\right|}{\left\|\boldsymbol{w}_{k}^{'}\right\|_{2}}$
            \STATE $\Delta \delta_{i} \leftarrow \frac{\left|f_{\hat{i}}^{'}\right|}{\left\|\boldsymbol{w}_{\hat{\imath}}^{'}\right\|_{2}^{2}} \boldsymbol{w}_{\hat{l}}^{'}$
            
             \STATE  Update the perturbation:
                $\delta\leftarrow \mathbbm{P}_{p,\epsilon}(\delta+\Delta \delta_i)$\\
              \ENDIF
             \ENDFOR
        \ENDWHILE
  \end{algorithmic}
\end{algorithm}

% \begin{figure}
%     \centering
%     \includegraphics[width=\linewidth]{figure/UAP.png}
%     \label{alg:UAP}
% \end{figure}

%Given a state-of-the-art deep neural network classifier, we show the existence of a universal (image-agnostic) and very small perturbation vector that causes natural images to be misclassified with high probability. We propose a systematic algorithm for computing universal perturbations, and show that state-of-the-art deep neural networks are highly vulnerable to such perturbations, albeit being quasi-imperceptible to the human eye. We further empirically analyze these universal perturbations and show, in particular, that they generalize very well across neural networks. The surprising existence of universal perturbations reveals important geometric correlations among the high-dimensional decision boundary of classifiers. It further outlines potential security breaches with the existence of single directions in the input space that adversaries can possibly exploit to break a classifier on most natural images.

\subsubsection{SPM}
%Art of Singular Vectors and Universal Adversarial Perturbations
Khrulkov and Oseledets~\cite{khrulkov2018art} proposed another method for computing the UAP by maximizing the difference between the activations of $i$-th hidden layer for a clean image $x$ and a perturbed image $x + \delta$. 
Mathematically, it can be formulated as follows:
\begin{equation}
\label{eq:spm}
    ||f_i(x+\delta) - f_i(x)||_q \rightarrow \max, \quad  ||\delta||_p = \eta.
\end{equation}
For a small perturbation vector $\delta$, this method utilizes the $(p,q)$-singular vectors of the Jacobian matrix of the features from the hidden layers so that the difference between the activations can be approximated by:
\begin{equation}
    f_i(x+\delta) - f_i(x) \approx J_i(x)\delta,
\end{equation}
where $J_i(x)=\frac{\partial f_i}{\partial x}|_x$ is the Jacobian matrix of $f_i$. This allows us to express Eq.~\ref{eq:spm} as follows:
\begin{equation}
\label{eq:spm2}
    ||f_i(x+\delta) - f_i(x)||_q \approx ||J_i(x)\delta||_q,
\end{equation}
To maximize the value on the left-hand side of Eq.~\ref{eq:spm2}, it is equivalent to maximize the right-hand side of Eq.~\ref{eq:spm2}. Therefore, Khrulkov and Oseledets~\cite{khrulkov2018art} construct the following loss function to compute the UAP,
\begin{equation}
    ||J_i(x)\delta||_q \rightarrow \max, \quad ||\delta||_p = \eta.
    \label{eq:spm3}
\end{equation}
where $\eta$ is a constant and usually set to 1 during the optimization. They use the generalized power method to compute the solution of the above Eq.~\ref{eq:spm3}.

%\subsection{Class-Discriminative UAPs}
\subsubsection{DT-UAP}
%Double Targeted Universal Adversarial Perturbations
Generally, the UAP can attack images from any category. However, in some cases, it is preferred to attack images from several selected categories to raise slight suspicion. Therefore, Benz et al. ~\cite{benz2020double} introduce the double-targeted universal adversarial perturbations (DT-UAPs). Specifically, the DT-UAPs only attack samples $X_s$ from one selected source targeted class $s$ to a given targeted class $t$, while having a limited adversarial effect on other samples $X_{ns}$ from those non-selected source classes. To achieve this goal, the DT-UAPs optimize the following loss function which simultaneously considers the selected class $X_s$ and the non-selected classes $X_{ns}$,
\begin{equation}
    L = L_t(X_s) + \alpha L_{ce}(X_{ns}).
\end{equation}
The first term $L_t(X_s)$ is related to the selected class and contains the following two parts:
\begin{align}
\label{eq:s1}
    L_{t}(X_s) &= \sum_{x_s} \max \Big(l_{c}(x_{s}+\delta)-\max _{i \neq c}\big(l_{i}\left(x_{s}+\delta\right)\big), 0 \Big) \\ \nonumber     
    &+\sum_{x_s} \max \Big(\max _{i \neq t}\left(l_{i}\left(x_{s}+\delta\right)-l_{\text t}\big(x_{s}+\delta\right)\big),-m\Big),
\end{align}
% \begin{equation}
%     L_{2}(X_s) = \sum_{x_s} \max \Big(\max _{i \neq t}\left(l_{i}\left(x_{s}+\delta\right)-l_{\text t}\big(x_{s}+\delta\right)\big),-m\Big),
% \end{equation}
where $l_{i}$ indicates the $i$-th entry in the predicted logits, $c =\arg \max(l(x_s))$, and $m$ is a margin. For samples from the non-selected classes, the $L_{ce}(X_{ns})$ computes the normal cross-entropy loss function for both original samples and their corresponding adversarial samples, enforcing the adversarial samples to maintain the original ground truth labels.

\subsubsection{CD-UAP} 
%Class Discriminative Universal Adversarial Perturbation}
The CD-UAP~\cite{zhang2020cd} extends one selected source targeted class in DT-UAP~\cite{benz2020double} to multiple selected source targeted classes. To craft the UAP, the CD-UAP optimizes the following loss function, 
\begin{equation}
    L = \alpha L_t(X_s) + \beta L_{ce}(X_{ns}),
\end{equation}
where $L_t(X_s)$ is an extension of Eq.~\ref{eq:s1} that computes the loss of samples from the multiple selected source targeted classes.

\subsubsection{FFF}
Mopuri et al.~\cite{mopuri2017fast} propose Fast Feature Fool (FFF), which is the first method that does not require source data, to craft the UAPs. The main motivation of FFF is that misfiring the activation of each feature layer in the CNNs can cause the model to produce wrong predictions. Therefore, the objective of FFF is to find the $\delta$ by maximizing the feature activation when feeding it to the target CNN model. The corresponding optimization loss function is 
\begin{equation}
    L_{FFF} = -\log\left( \prod_{i=1}^{K} \bar{f}^i(\delta) \right),
    \label{eq:fff}
\end{equation}
where $\bar{f}^i(\delta)$ is the mean feature activation of $\delta$ at the layer $i$, and $K$ is the number of convolutional layers in the target CNN. %In their experiment, the $\delta$ learned by the FFF shows a strong transferability to other CNN models.

\subsubsection{GD-UAP} %Generalizable Data-free Objective for Crafting Universal Adversarial Perturbations
Mopuri et al.~\cite{mopuri2018generalizable} further propose the GD-UAP, which incorporates the data-prior information into the previous FFF~\cite{mopuri2017fast} for crafting the UAP. The first prior is the mean and sigma computed from the training dataset, \textit{i.e.}, ($\mu, \sigma$). By leveraging the ($\mu, \sigma$), pseudo training sample $x$ can be sampled from a Gaussian distribution $\mathcal{N(\mu, \sigma)}$, and the loss function will become to,
\begin{equation}
    L_{FFF}^{\mu, \sigma} = -\sum_{x \sim \mathcal{N(\mu, \sigma)}} \log \left(\prod_{i=1}^{k} || f^i(x+\delta)||_2 \right).
\end{equation}

During the training, the authors observe that the $\delta$ will quickly surpass the imposed norm constraint $\epsilon$. Then, the GD-UAP additionally uses an improved optimization technique during the training, which re-scales the $\delta_i$ to its half after each iteration ($\delta_i \leftarrow \delta_i / 2$). Besides, the GD-UAP further exploits a subset $\mathcal{X}$ of real training data as prior information for crafting the UAP and optimizes the following loss function,
\begin{equation}
    L_{FFF}^{\mathcal{X}} = -\sum_{x \sim \mathcal{X}} \log \left(\prod_i^{k} || l_i(x+\delta)||_2 \right).
\end{equation}

\subsubsection{PD-UA}
%Universal Adversarial Perturbation via Prior Driven Uncertainty Approximation
The previous FFF~\cite{mopuri2017fast} and GD-UAP~\cite{mopuri2018generalizable} craft the UAP by maximizing the activation, which essentially rely on the uncertainty of a CNN model.  Liu et al.~\cite{liu2019universal} further propose a Prior-driven Uncertainty Approximation (PD-UA) method, which combines two types of uncertainty: Epistemic uncertainty and Aleatoric uncertainty. The former Epistemic uncertainty mainly reflects the number of credible activated neurons at each convolutional layer. The later Aleatoric uncertainty represents the stable quantity for various input data. During the implementation, the Epistemic uncertainty is achieved by maximizing the activated neurons at all convolutional layers through a Monte Carlo Dropout process~\cite{gal2016dropout}. The Aleatoric uncertainty leverages a texture bias as a prior regularizer to tune the distribution of the final perturbation $\delta$, which helps improve the attacking performance.

\subsubsection{F-UAP}
%: Understanding Adversarial Examples From the Mutual Influence of Images and Perturbations
To exploit the underlining property of the UAP, Zhang et al.~\cite{zhang2020understanding} analyze the influence of image and perturbation by computing their corresponding Person correlation coefficient (PCC). Based on the PCC, the authors find that the UAP in adversarial examples contains dominant features, while the original clean image is more like noise. Consequently, Zhang et al.~\cite{zhang2020understanding} propose the method F-UAP for generating targeted universal adversarial perturbations by using random proxy source images instead of the ImageNet training set. Besides, the authors propose new loss functions for the adversarial attack by enlarging the logit margin in both targeted and non-targeted attacks.

In the targeted attack, the loss function mainly aims at increasing the logit margin between the targeted class $t$ and other non-targeted classes, denoted as,
\begin{equation}
    L_t=\max\left( \max_{i\neq t} l_{i}(x+\delta) - l_{t}(x+\delta), -m\right),
\end{equation}
where $x$ are samples from a proxy dataset, $l_{i}$ is the $i$-th entry of the logit vector, and $t$ is the target class, $m$ is a hyper-parameter.

In the non-targeted attack, the F-UAP requires the source training dataset for crafting the UAPs. The corresponding margin-based loss function for the non-targeted attack is
\begin{equation}
    L_{nt} = \max\left( l_{y}(x+\delta) - \max_{i\neq y} l_{i}(x+\delta), -m\right).
\end{equation}
where $y$ are the ground-truth labels of $x$. 

% \textcolor{red}{For the subsequent experiments, we modified the number of data iterations, the overall optimization procedure is illustrated in Algorithm~\ref{alg:F-UAP}. }

\subsubsection{Jigsaw-UAP}
%: Data-Free Universal Adversarial Perturbation and Black-Box Attack
Jigsaw-UAP~\cite{zhang2021data} is a follow-up work of F-UAP~\cite{zhang2020understanding}, which first revisits the mechanism of the dominant label phenomenon in the non-targeted UAP. The authors observe that the estimated labels of most adversarial samples are usually identical. Besides, the average logits computed from all adversarial samples in the training set have a similar distribution with the logits of only using the UAP as input for the training CNN model. Based on these analyses, the authors aim to craft the UAPs by using artificial jigsaw images. To mimic the property of nature images and increase diversity, the randomly generated artificial jigsaw images follow two criteria: 1) locally smooth; 2) mixed frequency pattern. Moreover, a self-supervision cosine similarity loss function $L_{cos}$ is used to optimize the UAP by decreasing the logits' similarity,
\begin{equation}
    %L_{cos} = {CosSim}\left(l(x), l(x+\delta)\right),
    L_{cos} = \frac{l^T(x) l(x+\delta)}{||l(x)||_2||l(x+\delta)||_2},
\end{equation}
where $l(\cdot)$ are the output logits from the training CNN model.

\subsection{Generator-based attack methods}
\label{sec:methods_GM}
\begin{figure}
    \centering
    \includegraphics[width=0.9\linewidth]{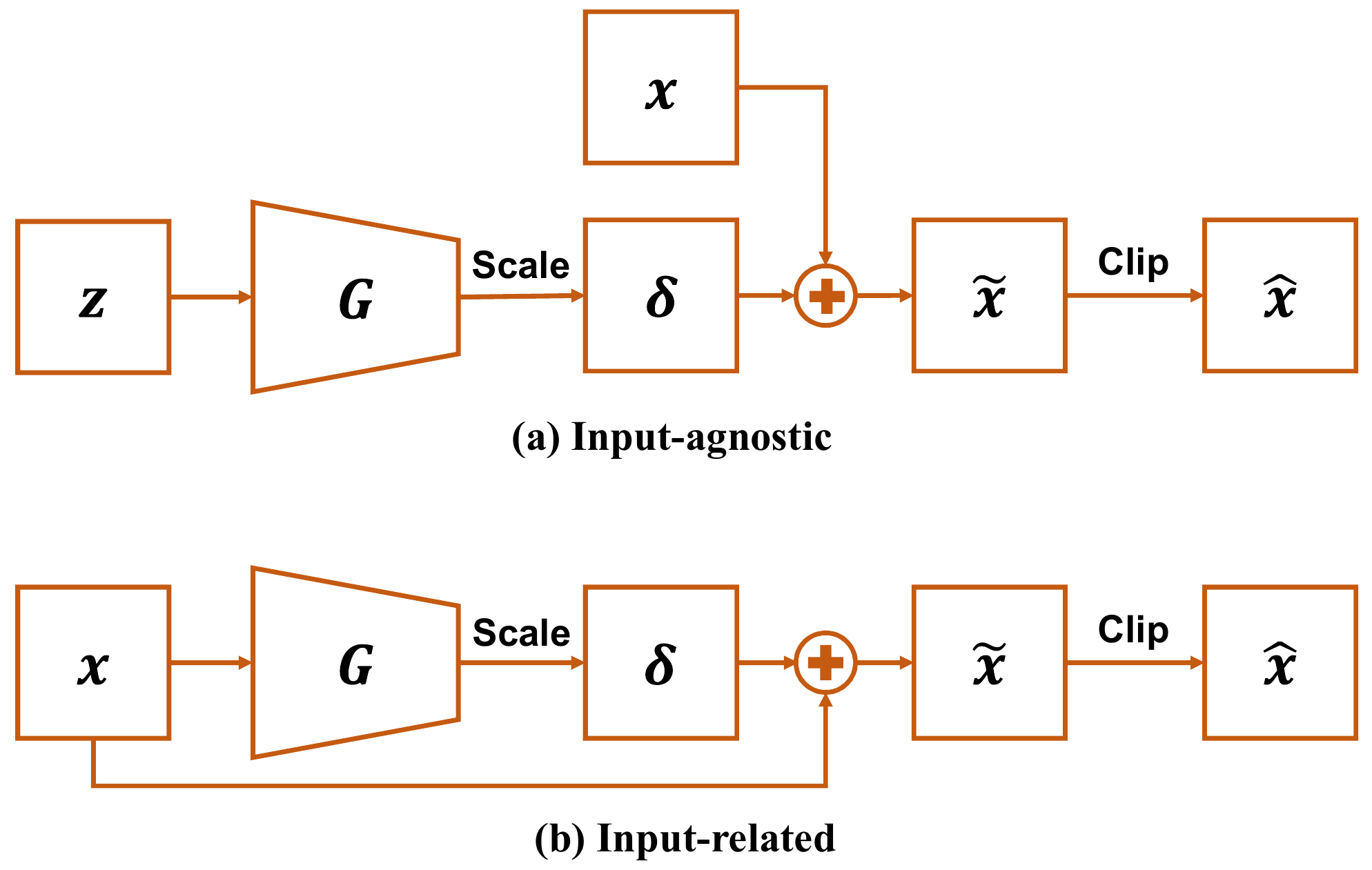}
    \caption{The overall pipelines of two different Generator-based methods for crafting the adversarial perturbations, \textit{i.e.}, input-agnostic and input-related.}
    \label{fig:gm_methods}
\end{figure}

In contrast with previous noise-based attack methods, the generator-based methods train an extra generative model $\bm{G}$ as a bridge to craft the perturbations indirectly. The generator-based methods can be mainly categorized into input-agnostic and input-related, as shown in Figure~\ref{fig:gm_methods}. In input-agnostic methods~\cite{poursaeed2018generative}, the generator $\bm{G}$ takes a random noise vector or a random noise image $\bm{z}$ as the input and computes the corresponding perturbation $\bm{\delta}$. A scale operation is performed to restrict $\bm{\delta}$ within the pre-defined maximum perturbation norm $\epsilon$. After the scaling operation, the $\bm{\delta}$ is added into $x$ to obtain the adversarial sample $\Tilde{x}$. A clip operation is used to set the $\Tilde{x}$ within the valid pixel value range.
On the other aspect, the input-related methods~\cite{hashemi2020transferable} take the original image $x$ as input and generate the input-related perturbation $\bm{\delta}$. Although the input-related methods utilize the original image $\bm{x}$ as input to generate specific perturbations $\bm{\delta}$ for each $\bm{x}$, we consider these methods as UAPs in this paper. This is because once the generator is trained, it is universal for any different input images.

Additionally, in generator-based methods, the training of the generative network (denoted as $G$) involves utilizing a targeted CNN network (denoted as $D$) to compute the adversarial loss functions. The original image $x$ can be the source data or proxy data, similar to noise-based methods. In the following sections, we will provide a detailed discussion of various generator-based methods.

% For image-agnostic setting, the GAP randomly samples a fixed pattern $z\in[0,1]^n$ from a uniform distribution $U[0,1]^n$ and feeds to a generator $G$ for creating the universal perturbation $\delta=G(z)$, which is then scaled to have a fixed norm. Finally, the perturbation $\delta$ is added to natural images to create the adversarial examples $\hat{x}$. While in image-dependent setting, this method uses a generator $G$ to generate perturbation for each natural image 
% this method seeks $\Theta$ such that for most images $x\in X:f(G(x)\neq f(x)$.   

\subsubsection{GAP}
%:Generative Adversarial Perturbations
Poursaeed et al.~\cite{poursaeed2018generative} firstly adopt the generative model for crafting Generative Adversarial Perturbations (GAP) in both non-targeted and targeted attacks. In~\cite{poursaeed2018generative}, the GAP evaluates both input-agnostic and input-dependent architecture.
The GAP adopts the cross-entropy loss with the least likely class for the non-targeted attack as follows,
\begin{equation}
    L_{nt} = L_{ce}(f(x+\delta),\mathbbm{1}_{ll}), 
\end{equation}
where $ll= \arg\min f(x)$ is the least likely class for the training sample $x$, $\mathbbm{1}_{ll}$ is the corresponding one-hot vector. Besides, the GAP also uses the negative CE loss for training non-targeted generator, depicted as:
\begin{equation}
    L_{nt} =- L_{ce}(f(x+\delta),\mathbbm{1}_{y}), 
\end{equation}
where $y$ is the ground-truth for image $x$.

In the targeted attack, the objective loss function is to minimize the cross-entropy with the targeted class, denoted as,
\begin{equation}
    L_{t}=L_{ce}(f(x+\delta),\mathbbm{1}_{t}).
\end{equation}

\subsubsection{GM-UAP}
%: Learning Universal Adversarial Perturbations with Generative Models
Inspired by the GAN, Hayes and Danezis\cite{hayes2018learning} leverages an input-agnostic generative network to craft the universal adversarial perturbation, namely, GM-UAP. The generator takes a random vector $\bm{z}$ sampled from a normal distribution $\mathcal{N}(0,1)^{100}$ as the input, and it outputs the corresponding perturbation $\bm{\delta}$. In the non-targeted attacking scenario, the GM-UAP~\cite{hayes2018learning} uses the following loss function for training,
\begin{equation}
\small
    L_{nt} =  \max \left\{ \log[f_c(\delta +x)] - \max_{i\neq c} \log[f_i(\delta +x)], -m \right\}
             + \alpha || \delta ||_p,
\end{equation}
where $c = \arg \max f(x)$ is the predicted class label of the input $x$, and $m$ is a confidence threshold. The second term $\alpha || \delta ||_p$ minimizes the norm of the UAP.

In the targeted attacking scenario, the following function is used to optimize the predicted label of the adversarial sample to the targeted label $t$,
\begin{equation}
\small
    L_{t} =  \max \left\{ \max_{i\neq t} \log[f_i(\delta +x)] -  \log[f_t(\delta +x)], -m \right\}
              + \alpha || \delta ||_p.
\end{equation}

\subsubsection{NAG}
%: Network for Adversary Generation
Similar to the GM-UAP~\cite{hayes2018learning}, NAG~\cite{mopuri2018nag} also uses an input-agnostic generator to craft the universal adversarial perturbation. In the NAG, the authors propose a novel loss function for training the generator and further use a diversity objective function to increase the fooling rate and the cross-model transferability. The overall computation flow of the NAG is as follows. 
Given a training batch with $N$ input images $\{x_1, ..., x_N\}$, it first samples $N$ random noise input vectors $\{z_1, ..., z_N\}$. The generator $G$ computes the corresponding perturbations $\{\delta_1, ..., \delta_N\}$.
After that, a shuffle operation is conducted to obtain the shuffled perturbations $\{\delta_{1^{'}}, ..., \delta_{N^{'}}\}$. 
% Then, the NAG optimizes the following loss functions.

For each clean input image $x_i$, we can obtain the predicted label $c=\arg\max f(x)$. %The desired $\delta$ should have the ability to change the prediction of $(x+\delta)$ different from $c$. 
Then, the NAG proposes the following fooling loss function to minimize the confidence of perturbed sample $(x+\delta)$ on the label $c$,
\begin{equation}
    L_{f} = -\log \left(1-f_{c}(x+\delta)\right).
\end{equation}

To further increase the diversity of generated perturbations, a diversity objective function is used to increase the distance of between two different adversarial samples of the same $x_n$. The diversity objective is computed by
\begin{equation}
    L_{d} = -\sum_{n=1}^{B} D\left(f^{i}(x_n + \delta_n), f^{i}(x_n + \delta_{n^{'}}) \right)
\end{equation}
where $f^i$ denotes the feature at layer-$i$ of the model $f$. 

The final loss function for training the perturbation generator $G$ is a combination of $L_{f}$ and $L_{d}$,
\begin{equation}
    L = L_{f} + \lambda L_{d},
    \label{loss:NAG}
\end{equation}
where $\lambda$ is a hyper-parameter.

\subsubsection{GM-TUAP}
%: Transferable Universal Adversarial Perturbations Using Generative Models
The GM-TUAP~\cite{hashemi2020transferable} is another input-agnostic generator-based method. In the GM-TUAP, the authors observe that the feature representations in the first layer of various model architectures share similar feature maps. Therefore, the GM-TUAP argues that increasing the adversarial energy in the first layer from a target model can improve the transferability of the generated perturbation for attacking unknown models. Therefore, the GM-TUAP incorporates the Fast Feature Fool (FFF)~\cite{mopuri2017fast} as an extra supervision for training. Specifically, the FFF is applied at the feature of the first layer of the generated adversarial sample $\hat{x}$. The corresponding loss function for the non-targeted scenario $L_{nt}$ and the target scenario $L_{t}$ are as follows, respectively. In the non-targeted, the loss function is
\begin{equation}
    L_{nt} = -\alpha L_{ce}\left( f(x + \delta), \mathbbm{1}_{y}\right) + (1-\alpha) L^1_{FFF}(x + \delta),
\end{equation}
where $\mathbbm{1}_{y}$ is the one-hot vector of the ground truth label of image $x$. $L^1_{FFF}$ denotes the Fast Feature Fool loss only at the first layer of the adversarial sample $x+\delta$.

For the targeted perturbation, the corresponding loss function is 
\begin{equation}
    L_{t} = -\alpha L_{ce}\left( f(x+\delta),  \mathbbm{1}_{t} \right) + (1-\alpha) L^1_{FFF}(x+\delta).
\end{equation}

\subsubsection{AAA}
%Ask, Acquire, and Attack: Data-free UAP Generation using Class Impressions
The previous GM-UAP~\cite{hayes2018learning}, NAG~\cite{mopuri2018nag}, GM-TUAP~\cite{hashemi2020transferable} methods need to use the actual source data for training, which are not applicable for the practical scenarios when the source data is not available. To deal with this issue, Mopuri et al.~\cite{mopuri2018ask} propose a two-stage training schema, ``Ask, Acquire and Attack (AAA)." In the first ``Ask and Acquire" stage, the target model is used to generate proxy samples (class impressions) through back-propagation by maximizing the label-wise prediction confidence. In the second ``Attack" stage, the generated proxy samples are then leveraged to train the perturbation generator by using the loss function (Eq.~\ref{loss:NAG}) proposed in NAG~\cite{mopuri2018nag}.

\subsubsection{CD-TAP}
%: Cross-Domain Transferability of Adversarial Perturbations
Naseer et al.~\cite{naseer2019cross} propose the input-related adversarial perturbation generator, namely CD-TAP. In the CD-TAP, the authors use a proxy source dataset for training to increase cross-domain transferability. Besides, a relative cross-entropy function approach is adopted to explicitly consider the logit outputs between the clean images and their adversarial counterparts. %Specifically, the discriminator will output low confidence scores for the perturbed adversarial images, while maintain the confidence scores of clean images. 
In the untargeted attack, the corresponding relative objective loss function is as follows,
\begin{equation}
    L_{rce} = - L_{ce}\left(f(x + \delta)-f(x), \mathbbm{1}_c \right),
\end{equation}
where $\mathbbm{1}_c$ is the one-hot label vector related to the estimate label $c=\arg\max(f(x))$ of input $x$. 

In the targeted attack, the corresponding relative objective loss function is as follows,
\begin{equation}
    L_{rce} = L_{ce}\left(f(x + \delta), \mathbbm{1}_{t} \right)+L_{ce}\left(f(x), \mathbbm{1}_c \right).
\end{equation}

\subsubsection{TTP}
%: On Generating Transferable Targeted Perturbations
The TTP~\cite{naseer2021generating} is a follow-up work of the CD-TAP, which focuses on the transferable targeted attack. Instead of maximizing the decision gap in CD-TAP, the TTP leverages a matching distribution loss for training, which minimizes the label distribution between adversarial samples of a proxy dataset and the original training source dataset. The matching distribution loss is implemented by the KL-divergence, denoted as,
\begin{equation}
    L_{md} = \sum_{i=1}^{N} f(x_s^i + \delta) \log \frac{f(x_s^i + \delta)}{f(x_t^i)} +  f(x_t^i)\log \frac{f(x_t^i)}{f(x_s^i + \delta))},
\end{equation}
where $x_s^i + \delta$ is an adversarial sample obtained from the proxy dataset, $x_t^i$ is a source training sample whose real ground-truth label is the target label $t$.

During the training, the TTP also adopts a data augmentation on $x_s$ to increase attacking performance based on the input transformations. Besides, a neighborhood similarity matching is further introduced to consider the local structure between the proxy dataset and the target domain. This neighborhood similarity matching loss $L_{sim}$ is also implemented by the KL-divergence with local normalization, 
\begin{equation}
    L_{sim} = \sum_{i,j} \bar{S}^t_{i,j} \log \frac{\bar{S}^t_{i,j}}{\bar{S}^s_{i,j}} + \sum_{i,j} \bar{S}^s_{i,j} \log \frac{\bar{S}^s_{i,j}}{\bar{S}^t_{i,j}}, 
\end{equation}
where ${S}^t_{i,j}$ computes the logits' similarity of two samples from the original target samples. ${S}^s_{i,j}$ computes the logits' similarity between a perturbed source domain sample and its augmented adversarial example in the proxy source dataset. $\bar{S}$ computes the softmax normalized similarity along the row dimension. By jointly considering the matching distribution loss and the neighborhood similarity matching loss, the final loss function for training TTP is 
\begin{equation}
    L_g = L_{md} + L_{md}^{aug} + L_{sim}.
\end{equation}

\begin{algorithm}[h!]
  \caption{Noise-based Framework. }
  \label{alg:F-UAP}
  \textbf{Inputs:} Training data $X$, Loss function $L$, the perturbation budget $\epsilon$ under the $l_p$ norm, epoch $N$. \\
  \textbf{Outputs:} Universal perturbation $\delta$. \\
  \begin{algorithmic}[1]
    \STATE Initialize $\delta\leftarrow 0$.\\
    \FOR{$epoch=1,\cdots,N$} 
    \FOR{ $B$ in  $X$ }
    \STATE $ g_{\delta} \leftarrow \underset{x \sim B}{\mathbb{E}}\left[\nabla_{\delta} L\right]  \quad \triangleright \text { Calculate gradient }$ \\
    \STATE $\delta \leftarrow \text {optim}\left(g_{\delta}\right)  \quad \triangleright \text { Update } $  \\
    \STATE $\delta  \leftarrow \epsilon \frac{\delta}{\|\delta\|_p} \quad \quad \quad \triangleright \text { Norm projection } $ 
    \ENDFOR
    \ENDFOR  
  \end{algorithmic}
\end{algorithm}

\begin{algorithm}[h!]
  \caption{Generator-based Framework.  }
  \label{alg:GAP}
  \textbf{Inputs:} Training data $X$, Loss function $L$, the perturbation budget $\epsilon$ under the $l_p$ norm, epoch $N$. \\
  \textbf{Outputs:} A generative network $G$. \\
  \begin{algorithmic}[1]
    \STATE Initialize weights of $G$.\\
    \IF{ Input-agnostic}
    \STATE Random noise image (vector) $z$.\\
    \ENDIF
    \FOR{$epoch=1,\cdots,N$} 
    \FOR{ $B$ in  $X$ }
    \IF{ Input-related } 
    \STATE $ \delta_m = G(B) \quad \triangleright \text {Craft perturbations}$ \\

    \ELSIF{ Input-agnostic}
    \STATE $ \delta_m = G(z) \quad \triangleright \text {Craft perturbations}$ \\
    \ENDIF
    \STATE $\text {NormScale}(\delta_m) \quad \triangleright \text { Norm projection } $ \\
    \STATE $\tilde{B}=\delta_m + B  \quad \triangleright \text {Craft Adversarial examples } $\\ 
    \STATE  $ g_{G} \leftarrow \underset{x \sim \tilde{B}}{\mathbb{E}}\left[\nabla_{G} L\right] \quad \triangleright \text { Calculate gradient } $\\
        \STATE $G \leftarrow \text {optim}\left(g_{G}\right) \quad \triangleright \text { Update } $  \\
    \ENDFOR
    \ENDFOR
         
  \end{algorithmic}
\end{algorithm}

\begin{table*}[t]
\centering
\caption{The common loss functions used for training the UAPs. *$c= \arg\max f(x)$ and $ll= \arg\min f(x)$}
\label{tab:loss_fn}
% \resizebox{1\linewidth}{!}{
\begin{tabular}{c|c|l}
\hline
\textbf{Type} & \textbf{Name} & \makecell[c]{\textbf{Equation}}    \\ \hline
Feature-based & FFF & $ L_{FFF} = -\sum_{x \sim \mathcal{X}} \log \left(\prod_i^{k} || f^i(x+\delta)||_2 \right)$ \\
\hline
Similarity-based & Cosine & $L_{cos} = \frac{l^T(x) l(x+\delta)}{||l(x)||_2||l(x+\delta)||_2}$\\ \hline
\multirow{5}{*}{Entropy-based} 
& Targeted $CE$ & $L_{ce}=L_{ce}(f(x+\delta),\mathbbm{1}_{t})$ \\
& Targeted Relative $CE$ & $L_{rce}^{t} = L_{ce}\left(f(x+\delta), \mathbbm{1}_{t}  \right)+L_{ce}\left(f(x), \mathbbm{1}_c \right)$ \\
& Non-Targeted Relative $CE$ & $L_{rce}^{nt} = - L_{ce}\left(f(x+\delta)-f(x), \mathbbm{1}_c \right)$\\ 
& Negative $CE$ & $L_{nce}= -L_{ce}(f(x+\delta),\mathbbm{1}_{c})$\\
& Least Likely $CE$ & $L_{cell}= L_{ce}(f(x+\delta),\mathbbm{1}_{ll})$\\
\hline
\multirow{2}{*}{Margin-based}  
& $C\&W_{t}$ & $L_{cw}^{t}=\max\left( \max_{i\neq t} l_{i}(x+\delta) - l_{t}(x+\delta), -m\right) $\\
& $C\&W_{nt}$ & $ L_{cw}^{nt} = \max\left( l_{c}(x+\delta) - \max_{i\neq y} l_{i}(x+\delta), -m\right)$  \\
\hline
\end{tabular}
% }
\end{table*}

\subsubsection{GAP++}
%Learning to generate target-conditioned adversarial examples
Inspired by the conditional GAN, Mao et al.~\cite{mao2020gap++} propose the GAP++, which takes the original image as input and further treats the target label as a latent conditional vector to craft the conditional perturbations. In the GAP++, the latent vector is a one-hot vector with targeted-class $t$ in the targeted attack and is all zeros in the untargeted attack. Finally, the generator is trained by minimizing the cross-entropy between the output probability of adversarial samples and the latent vector. By doing so, the GAP++ can be used to craft both targeted and untargeted adversarial perturbations.

\subsection{Summary of the loss functions}
In the previous sections, we discussed the computation details of both noise-based and generator-based methods.
We can observe that different training frameworks adopt various loss functions for learning the universal adversarial perturbation. And these loss functions can be broadly categorized into the following groups: feature-based, similarity-based, entropy-based, and margin-based. 

The representative loss functions are listed in Table~\ref{tab:loss_fn}. 
\textbf{1)} Feature-based loss functions primarily aim to disrupt the activation of features in the hidden layers of the white-box model. 
% These loss functions focus on minimizing the similarity or distance between the features of the original sample and its corresponding adversarial example.
\textbf{2)} Similarity-based loss functions are designed to decrease the similarity between the logits of the original sample and its adversarial counterpart. 
% The objective is to create an adversarial example that is classified differently from the original sample.
\textbf{3)} Entropy-based loss functions are proposed to manipulate the final classification probability of the adversarial example. In non-targeted settings, negative cross-entropy (CE) and relative cross-entropy are employed to guide the perturbed sample toward the nearest decision boundary, while the least likely cross-entropy encourages the sample to be classified into a semantically distant class. In targeted settings, cross-entropy and relative cross-entropy are used to steer the sample toward the decision boundary of the target class.
\textbf{4)} Margin-based loss functions aim to increase the logits margin. This moves the adversarial example towards a class that is semantically similar to the original class but far away from the ground-truth class. In targeted attacks, the objective is to move the adversarial example towards the target class, which is far away from the ground-truth class. 

In this survey, we aim to address the lack of a systematic comparison of loss functions under the same training frameworks, including
noise-based and generator-based. To gain a deeper understanding of the characteristics of different loss functions in UAPs, we conduct comprehensive large-scale experiments to evaluate their attacking performance.

\begin{table*}[ht!]
\caption{The fooling rates (\%) of different non-targeted attacks in white-box setting.}
\label{tab:non-white}
\centering
\begin{tabular}{c|c|cccc|cccc} \hline
                              &                        & \multicolumn{4}{c|}{Noise-based}           & \multicolumn{4}{c}{Generator-based}                                                                               \\
\multirow{-2}{*}{Training Data} & \multirow{-2}{*}{Loss} & VGG-16 & VGG-19 & ResNet-50 & AVG   & VGG-16 & VGG-19 & ResNet-50 & AVG   \\ \hline
                              & $L_{fff}$   & 95.81  & 95.63  & 78.48  & 89.97  & 96.69  & 94.17  & 85.69  & 92.18 \\
                              & $L_{cos}$   & 97.31  & 96.27  & 94.75  & 96.11  & 98.75  & 98.64  & 89.37  & 95.59 \\
                              & $L_{cw}$    & 88.75  & 88.68  & 87.32  & 88.25  & 98.97  & 98.54  & 96.23  & 97.91 \\
                              & $L_{nce}$   & 96.17  & 94.69  & 91.77  & 94.21  & 99.57  & 99.29  & 97.98  & 98.95 \\
                              & $L_{cell}$  & 89.60  & 89.39  & 84.80  & 87.93  & 91.59  & 95.30  & 76.42  & 87.77 \\
\multirow{-6}{*}{ImageNet}    & $L_{rce}$   & 96.15  & 94.07  & 92.39  & 94.20  & 99.57  & 99.54  & 96.37   & 98.49  \\ \hline
                              & $L_{fff}$   & 95.74  & 95.64  & 78.69  & 90.02  & 97.43  & 95.05  & 87.63  & 93.37 \\
                              & $L_{cos}$   & 96.95  & 96.44  & 93.29  & 95.56  & 98.19  & 98.08  & 91.93  & 96.07 \\
                              & $L_{cw}$    & 79.25  & 80.34  & 72.91  & 77.50  & 98.15  & 98.28  & 93.73  & 96.72 \\
                              & $L_{nce}$   & 94.62  & 94.62  & 90.55  & 93.27  & 99.20  & 99.41  & 97.97  & 98.86 \\
                              & $L_{cell}$  & 88.22  & 86.67  & 81.43  & 85.44  & 92.60  & 94.10  & 70.91  & 85.87 \\
\multirow{-6}{*}{COCO}        & $L_{rce}$   & 96.28  & 94.49  & 91.10  & 93.96  & 99.17  & 99.41  & 88.96  & 95.85 \\ \hline
\end{tabular}
\end{table*}

\section{Experiments}
\label{sec:experiment}
In this section, we perform a series of experiments to evaluate different loss functions under the same training frameworks, including noise-based and generator-based. 
For the noise-based framework, we update the UAP with
the calculated gradient proposed in representative and simple F-UAP~\cite{zhang2020understanding}, and the overall optimization procedure is illustrated in Algorithm~\ref{alg:F-UAP}. In the generator-based framework, we update the UAP generation network with conventional training techniques in GAP~\cite{poursaeed2018generative}, and the overall optimization procedure is shown in Algorithm~\ref{alg:GAP}. 
Since the input-related architecture is more commonly used, we adopt it as the default choice for our generator-based framework. Additionally, the loss functions $L$ in Algorithm~\ref{alg:F-UAP} and Algorithm~\ref{alg:GAP} can be chosen from any of the options listed in Table~\ref{tab:loss_fn}. The training data can be either the original data or proxy data. Our objective is to gain a better understanding of the strengths and limitations of different training frameworks and loss functions.

First, we compare two types of attackers across six loss functions for non-targeted attacks in white-box and black-box settings. Then, we evaluate three loss functions for targeted attacks in white-box and black-box settings. Finally, we evaluate these attackers against various unknown defense mechanisms. We generate adversarial examples by these methods for fooling CNNs pre-trained on the ImageNet dataset~\cite{deng2009imagenet}. 

\subsection{Experimental Setup}
We train universal attackers from $20k$ images of two datasets, the ImageNet~\cite{deng2009imagenet} training set and the proxy dataset, \textit{i.e.}, MS-COCO~\cite{lin2014microsoft}. Then, these trained universal attackers are used to attack the ImageNet validation set in all of the experiments. The white-box models include VGG-16~\cite{simonyan2014very}, VGG-19~\cite{simonyan2014very} and ResNet-50~\cite{he2016deep}. To evaluate transferability, we choose ResNet-18~\cite{he2016deep}, DenseNet-121~\cite{huang2017densely}, and GoogleNet~\cite{szegedy2015going} as black-box models. The hyper-parameters used for training are as follows: each attacker is trained for 5 epochs, the batch size is 20, and the perturbation magnitude is $\epsilon=10$.  Besides, we use Adam optimizer~\cite{kingma2014adam} during the training process. In the noise-based framework, we adopt a learning rate of $lr = 0.005$, following the approach used in F-UAP~\cite{zhang2020understanding}. In the generator-based framework,  we set the learning rate to $lr = 0.002$ following~\cite{poursaeed2018generative}.

\subsection{Results in Non-targeted Attacks}
In this part, we evaluate the performance of different loss functions in non-targeted attacks. The fooling rates of white-box attacks and black-box attacks are reported in Tab.~\ref{tab:non-white} and Tab.~\ref{tab:nt-black}, respectively. 

\subsubsection{White-box attacks}
\textbf{Firstly}, it is evident that both noise-based and generator-based frameworks can achieve high fooling rates under different loss functions. However, the overall fooling rate of the noise-based framework is slightly lower than that of the generator-based framework. Nevertheless, the generator-based framework does not exhibit superior performance compared to the noise-based algorithm in the white-box setting.
\textbf{Secondly}, when comparing the fooling rates obtained by training on the source ImageNet dataset versus the proxy COCO dataset, there is no significant difference between them. Therefore, using the original training source dataset is unnecessary for crafting the UAP.
\textbf{Thirdly}, it is observed that the fooling rate of ResNet-50 is lower than that of VGG-16 and VGG-19, indicating that ResNet-50 is slightly more robust to adversarial attacks.

\textbf{Finally}, analyzing the performance of using different loss functions for training yields the following observations. \textbf{1)} The $L_{cos}$, $L_{nce}$ and $L_{rce}$ can obtain a stable attack fooling rate ($>90\%$) in both noise-based and generator-based frameworks, either using ImageNet or COCO for training. 
\textbf{2)} For the feature-based $L_{fff}$, we can find that the performance of VGG-16 and VGG-19 is superior to that of ResNet-50. We argue the main reason for this phenomenon is that ResNet-50 contains the residual connection. 
\textbf{3)} For the margin-based $L_{cw}$, there is a large gap between the noise-based and generator-based frameworks in all three networks.
\textbf{4)} Comparing the entropy-based $L_{cell}$ with other two entropy-based loss functions $L_{nce}$ and $L_{rce}$, the fooling rate of $L_{cell}$ is around 6-8\% lower in the noise-based framework and 10-13\% in the generator-based framework. These results suggest that encouraging adversarial examples to find the furthest decision boundary in $L_{cell}$ is less effective than seeking the nearest decision boundary in $L_{nce}$ and $L_{rce}$.

\begin{table*}[t]
\caption{The fooling rates (\%) of different non-targeted attacks in the black-box setting.}
\label{tab:nt-black}
\begin{subtable}[t]{\linewidth}
\centering
\resizebox{0.95\linewidth}{!}{
\begin{tabular}{c|c|ccccc|ccccc}
\hline
\multirow{2}{*}{Data} & \multirow{2}{*}{Loss} & \multicolumn{5}{c|}{ResNet-50}                        & \multicolumn{5}{c}{VGG-19}                           \\
                            &                       & DenseNet-121 & GoogleNet & ResNet-18 & VGG-16 & AVG   & DenseNet-121 & GoogleNet & ResNet-18 & VGG-16 & AVG   \\ \hline
\multirow{6}{*}{ImageNet}        & $L_{fff}$    & 29.98  & 24.98  & 41.08  & 44.31  & 35.09  & 27.50  & 26.24  & 36.68  & 90.77  & 45.30 \\                      
                                 & $L_{cos}$    & 69.73  & 57.54  & 77.08  & 80.94  & 71.32  & 50.21  & 52.48  & 61.06  & 92.52  & 64.07 \\
                                 & $L_{cw}$     & 51.48  & 45.98  & 63.34  & 67.32  & 57.03  & 28.12  & 30.61  & 40.38  & 75.25  & 43.59 \\
                                 & $L_{nce}$    & 57.48  & 48.11  & 63.24  & 70.88  & 59.93  & 40.06  & 36.74  & 46.88  & 87.48  & 52.79 \\                                 
                                 & $L_{cell}$   & 51.88  & 48.75  & 62.31  & 66.12  & 57.27  & 39.66  & 39.52  & 49.21  & 79.44  & 51.96 \\
                                 & $L_{rce}$    & 58.75  & 49.33  & 65.56  & 72.46  & 61.53  & 37.94  & 35.35  & 45.26  & 84.33  & 50.72 \\
\hline
\multirow{6}{*}{COCO}            & $L_{fff}$    & 28.62  & 24.30  & 39.12  & 41.70  & 33.44  & 27.45  & 25.75  & 37.57  & 90.77  & 45.39 \\
                                 & $L_{cos}$    & 65.65  & 57.11  & 74.45  & 78.26  & 68.87  & 52.80  & 55.85  & 65.00  & 92.78  & 66.61 \\
                                 & $L_{cw}$     & 38.13  & 34.07  & 50.22  & 54.60  & 44.26  & 23.58  & 24.48  & 33.92  & 60.58  & 35.64 \\
                                 & $L_{nce}$    & 52.79  & 44.24  & 59.49  & 67.50  & 56.01  & 39.92  & 36.75  & 47.15  & 86.73  & 52.64 \\
                                 & $L_{cell}$   & 49.70  & 46.61  & 57.79  & 62.07  & 54.04  & 35.44  & 36.71  & 45.34  & 74.78  & 48.07 \\
                                 & $L_{rce}$    & 56.37  & 46.38  & 62.33  & 62.33  & 56.85  & 36.16  & 36.12  & 44.22  & 85.53  & 50.51 \\
\hline
\end{tabular}}
\caption{Noise-based framework}
\end{subtable}

\begin{subtable}[t]{\linewidth}
\centering
\resizebox{0.95\linewidth}{!}{
\begin{tabular}{c|c|ccccc|ccccc}
\hline
    &                        & \multicolumn{5}{c|}{ResNet-50}   &\multicolumn{5}{c}{VGG-19}                       \\
\multirow{-2}{*}{Data} & \multirow{-2}{*}{Loss}  & DenseNet-121  & GoogleNet  & ResNet-18  & VGG-16  & AVG  & DenseNet-121  & GoogleNet  & ResNet-18  & VGG-16  & AVG \\ \hline
                                  & $L_{fff}$   & 71.80  & 65.33  & 79.54  & 78.06  & 73.68  & 33.47  & 27.98  & 36.45  & 91.21  & 47.28 \\
                                  & $L_{cos}$   & 73.51  & 56.38  & 77.19  & 86.25  & 73.33  & 46.80  & 32.86  & 49.97  & 97.39  & 56.76 \\
                                  & $L_{cw}$    & 80.12  & 58.35  & 80.91  & 84.80  & 76.05  & 34.82  & 27.90  & 37.51  & 96.84  & 49.27 \\
                                  & $L_{nce}$   & 83.83  & 58.79  & 81.93  & 85.95  & 77.63  & 49.44  & 42.06  & 54.64  & 97.13  & 60.82 \\
                                  & $L_{cell}$  & 62.00  & 57.53  & 76.11  & 86.41  & 70.51  & 46.37  & 42.76  & 53.65  & 90.87  & 58.41 \\
\multirow{-6}{*}{ImageNet}        & $L_{rce}$   & 73.91   & 50.72   & 82.11   & 89.50  & 74.06   & 48.51  & 39.94  & 54.44  & 97.65  & 60.14 \\ \hline
                                  & $L_{fff}$   & 70.10  & 52.08  & 74.78  & 76.54  & 68.38  & 34.11  & 30.55  & 36.76  & 90.20  & 47.91 \\
                                  & $L_{cos}$   & 73.87  & 57.15  & 77.14  & 87.23  & 73.85  & 46.45  & 34.68  & 49.90  & 96.11  & 56.79 \\
                                  & $L_{cw}$    & 75.77  & 52.61  & 76.46  & 81.49  & 71.58  & 37.00  & 32.58  & 39.19  & 95.88  & 51.16 \\
                                  & $L_{nce}$   & 83.72  & 60.39  & 83.09  & 83.83  & 77.76  & 48.50  & 40.56  & 52.73  & 97.38  & 59.79 \\
                                 & $L_{cell}$  & 57.38  & 52.14  & 70.77  & 79.54  & 64.96  & 48.98  & 43.83  & 56.18  & 91.68  & 60.17 \\
\multirow{-6}{*}{COCO}            & $L_{rce}$   & 69.89  & 43.60  & 68.14  & 84.48  & 66.53  & 45.93  & 36.44  & 52.29  & 97.61  & 58.07 \\
\hline
\end{tabular}}
\caption{Generator-based framework}
\end{subtable}
\end{table*}

\begin{table*}[t]
\caption{White-box performance (tFR and ntFR in \%) of different universal attackers under 8-Targets setting.}
\label{tab:target_white}
\centering
\centering
\resizebox{0.95\linewidth}{!}{
\begin{tabular}{c|c|cccccccc|cccccccc}
\hline
\multirow{3}{*}{Data} & \multirow{3}{*}{LOSS} & \multicolumn{8}{c|}{Noise-based framework}                                                                                       & \multicolumn{8}{c}{Generator-based framework}                                                                                          \\
                      &  & \multicolumn{2}{c}{ResNet-50} & \multicolumn{2}{c}{VGG-16} & \multicolumn{2}{c}{VGG-19} & \multicolumn{2}{c|}{AVG} &  \multicolumn{2}{c}{ResNet-50} & \multicolumn{2}{c}{VGG-16} & \multicolumn{2}{c}{VGG-19} & \multicolumn{2}{c}{AVG} \\
                      &  & tFR  & ntFR  & tFR  & ntFR  & tFR  & ntFR  & tFR  & ntFR  & tFR  & ntFR  & tFR  & ntFR  & tFR  & ntFR  & tFR  & ntFR  \\ \hline
\multirow{3}{*}{ImageNet}      & $L_{cw}$  & 76.97  & 86.69  & 77.87  & 93.42  & 79.44  & 93.14  & 78.09  & 91.08   & 73.18  & 86.39  & 74.28  & 98.11  & 62.64  & 97.01  & 70.04  & 93.84 \\
                              & $L_{ce}$  & 73.17  & 86.07  & 78.81  & 89.91  & 77.81  & 91.30  & 76.59  & 89.10    &54.74  & 79.75  & 79.73  & 95.11  & 72.51  & 93.96  & 69.00  & 89.61\\
                              & $L_{rce}$ & 73.67  & 86.25  & 78.64  & 91.38  & 77.64  & 91.18  & 76.65  & 89.60   & 58.94  & 80.87  & 80.81  & 95.23  & 76.01  & 94.36  & 71.92  & 90.15\\ \hline 
\multirow{3}{*}{COCO}          & $L_{cw}$   & 76.03  & 85.69  & 77.69  & 93.24  & 81.38  & 93.09  & 78.37  & 90.67  & 72.61  & 85.37  & 70.60  & 98.13  & 67.36  & 96.97  & 70.19  & 93.49 \\
                              & $L_{ce}$  & 70.33  & 81.90  & 76.82  & 89.08  & 76.73  & 88.84  & 74.63  & 86.61    & 61.18  & 80.16  & 80.31  & 94.80  & 78.25  & 93.83  & 73.25  & 89.59 \\
                              & $L_{rce}$ & 70.57  & 81.88  & 76.51  & 87.58  & 76.30  & 88.68  & 74.46  & 86.05 & 60.74  & 80.07  & 80.52  & 94.79  & 78.85  & 93.90  & 73.37  & 89.59 \\ \hline
\end{tabular}}
\end{table*}

\subsubsection{Black-box Attacks}
Since the fooling rates of VGG-16 and VGG-19 are similar in white-box settings, we will focus on reporting the black-box transferability using ResNet-50 and VGG-19 as surrogate models. We consider DenseNet-121, GoogleNet, ResNet-18, and VGG-16 as the target black-box models. Based on the findings presented in Table~\ref{tab:nt-black}, we can draw the following conclusions: \textbf{1)} Comparing the results of the noise-based and generator-based frameworks, we observe that perturbations generated by the generator-based framework exhibit higher transferability when attacking unknown black-box models.
\textbf{2)} In most cases, training with the source ImageNet dataset yields slightly higher transferability compared to using the proxy COCO dataset.
\textbf{3)} When attacking DenseNet-121, GoogleNet, and ResNet-18, adversarial samples crafted using ResNet-50 demonstrate a higher fooling rate than those generated using VGG-19. However, VGG-19 exhibits strong transferability for attacking VGG-16, as they have similar architectures.
\textbf{4)} Analyzing the performance of different loss functions, we find that $L_{cos}$ performs best for the noise-based framework, while $L_{nce}$ yields superior results for the generator-based framework. Additionally, $L_{cos}$ shows promising transfer fooling rates in the generator-based framework. Moreover, the perturbations learned using $L_{fff}$ in the generator-based framework and ResNet-50 demonstrate higher transferability in black-box attacks.
\textbf{5)} Lastly, we observe that transferring from VGG-19 or ResNet-50 models to the GoogleNet model is relatively difficult in all attacks. This observation highlights the need to design robust models using the inception module proposed in GoogleNet.

\begin{table*}[t]
\centering
\caption{Black-box performance (tFR and ntFR in \%) of different universal attackers under 8-Targets setting.}
\label{tab:target_black}
\begin{subtable}[t]{\linewidth}
\centering
\resizebox{0.95\linewidth}{!}{
\begin{tabular}{c|c|cccccccc|cccccccc}
\hline
\multirow{3}{*}{Data} & \multirow{3}{*}{LOSS} & \multicolumn{8}{c|}{ResNet-50}                                                                                       & \multicolumn{8}{c}{VGG-19}                                                                                          \\
                      &  & \multicolumn{2}{c}{DenseNet-121} & \multicolumn{2}{c}{Googlenet} & \multicolumn{2}{c}{ResNet-18} & \multicolumn{2}{c|}{AVG} & \multicolumn{2}{c}{DenseNet-121} & \multicolumn{2}{c}{Googlenet} & \multicolumn{2}{c}{ResNet-18} & \multicolumn{2}{c}{AVG} \\
                      &  & tFR  & ntFR  & tFR  & ntFR  & tFR  & ntFR  & tFR  & ntFR  & tFR  & ntFR  & tFR  & ntFR  & tFR  & ntFR  & tFR  & ntFR  \\ \hline
\multirow{3}{*}{ImageNet}  & $L_{cw}$  & 18.35  & 48.81  & 2.64  & 39.27  & 14.41  & 54.98  & 11.80  & 47.69  & 1.01  & 37.79  & 0.25  & 35.87  & 1.20  & 44.71  & 0.82  & 39.46 \\
                           & $L_{ce}$  & 13.29  & 47.28  & 2.62  & 40.68  & 9.76   & 55.94  & 8.55   & 47.97  & 0.80  & 35.78  & 0.32  & 35.13  & 0.86  & 44.99  & 0.66  & 38.63 \\
                           & $L_{rce}$ & 12.94  & 43.39  & 2.08  & 40.63  & 8.81   & 55.56  & 7.94   & 46.53  & 0.72  & 36.01  & 0.24  & 35.92  & 0.88  & 45.14  & 0.61  & 39.02 \\ \hline
\multirow{3}{*}{COCO} & $L_{cw}$  & 17.94  & 48.13  & 8.53  & 33.42  & 14.29  & 54.22  & 13.59  & 45.26  & 0.81  & 37.66  & 0.22  & 36.21  & 0.88  & 44.89  & 0.64  & 39.59  \\
                      & $L_{ce}$  & 12.01  & 42.63  & 2.96  & 35.72  & 10.05  & 51.35  & 8.34   & 43.23  & 0.94  & 31.20  & 0.25  & 30.91  & 0.94  & 40.00  & 0.71  & 34.03  \\
                      & $L_{rce}$ & 12.51  & 42.33  & 2.83  & 35.57  & 10.82  & 51.34  & 8.72   & 43.08  & 1.05  & 31.28  & 0.30  & 30.71  & 0.78  & 39.42  & 0.71  & 33.80  \\ \hline
\end{tabular}}
\caption{Noise-based framework}
\end{subtable}
\begin{subtable}[t]{\linewidth}
\centering
\resizebox{0.95\linewidth}{!}{
\begin{tabular}{c|c|cccccccc|cccccccc}
\hline
\multirow{3}{*}{Data} & \multirow{3}{*}{LOSS} & \multicolumn{8}{c|}{ResNet-50}                                                                                       & \multicolumn{8}{c}{VGG-19}                                                                                          \\
                      &                       & \multicolumn{2}{c}{DenseNet-121} & \multicolumn{2}{c}{GoogleNet} & \multicolumn{2}{c}{ResNet-18} & \multicolumn{2}{c|}{AVG} & \multicolumn{2}{c}{DenseNet-121} & \multicolumn{2}{c}{GoogleNet} & \multicolumn{2}{c}{ResNet-18} & \multicolumn{2}{c}{AVG} \\
                      &                       & tFR           & ntFR         & tFR           & ntFR          & tFR         & ntFR        & tFR        & ntFR       & tFR          & ntFR          & tFR           & ntFR          & tFR         & ntFR        & tFR        & ntFR       \\ \hline
\multirow{3}{*}{ImageNet}  & $L_{cw}$  & 29.14  & 63.84  & 7.28  & 49.04  & 28.10  & 70.89  & 21.51  & 61.26  & 0.82  & 44.14  & 0.29   & 36.32   & 1.02  & 48.30  & 0.71  & 42.92      \\
                      & $L_{ce}$  & 17.78  & 62.29  & 4.80  & 49.47  & 15.65  & 67.95  & 12.74  & 59.90  & 1.94   & 44.20  & 0.59  & 38.25  & 2.10  & 50.07  & 1.55  & 44.17      \\
                      & $L_{rce}$  & 19.95  & 61.66  & 5.11  & 51.04  & 18.54  & 69.33  & 14.53  & 60.68  & 2.18  & 44.73  & 0.51  & 37.97  & 1.72  & 50.99  & 1.47  & 44.57    \\ \hline 
\multirow{3}{*}{COCO} & $L_{cw}$  & 25.35  & 61.82  & 6.77  & 48.29  & 24.99  & 69.21  & 19.04  & 59.77  & 1.04  & 41.17  & 0.22  & 34.16  & 0.84  & 45.74  & 0.70  & 40.35      \\
                      & $L_{ce}$  & 19.29  & 60.83  & 5.19  & 49.33  & 16.64  & 66.94  & 13.71  & 59.03  & 3.18  & 45.02  & 0.43  & 37.74  & 2.39  & 50.97  & 2.00  & 44.58      \\
                      & $L_{rce}$ & 19.15  & 60.61  & 5.38  & 50.63  & 17.99  & 67.96  & 14.18  & 59.73  & 1.83  & 43.58  & 0.32  & 37.53  & 2.12  & 52.00  & 1.42  & 44.37      \\ \hline
\end{tabular}
}
\caption{Generator-based framework}
\end{subtable}
\end{table*}

\begin{table*}[t]
\caption{The fooling rates (\%) of different non-targeted attacks against different defense mechanisms.}
\label{tab:nt-defense}
\begin{subtable}[t]{\linewidth}
\centering
\begin{tabular}{c|c|ccccccccc}
\hline
\multirow{2}{*}{Data} & \multirow{2}{*}{Loss} & Input Processing & \multicolumn{4}{c}{Adversarial Training}                        & \multicolumn{2}{c}{Stylized} & \multirow{2}{*}{Augmix} \\
                               &                       & NRP              & $L_{2}=0.1$ & $L_{2}=0.5$ & $L_{\infty} =0.5$ & $L_{\infty} =1$ & SIN          & SIN-IN        &                                 \\
\hline
\multirow{6}{*}{ImageNet} & $L_{fff}$  & 46.47 & 22.76       & 16.08       & 14.34             & 10.48           & 26.01 & 28.84  & 41.19  \\
                          & $L_{cos}$  & 23.19 & 24.86       & 17.18       & 16.74             & 12.75           & 45.14 & 74.43  & 63.52 \\
                          & $L_{cw}$   & 27.85 & 19.21       & 14.48       & 13.61             & 10.50           & 37.97 & 55.98  & 49.82 \\
                          & $L_{nce}$  & 17.91 & 24.47       & 17.24       & 16.20             & 12.62           & 41.15 & 66.18  & 52.62  \\
                          & $L_{cell}$ & 24.08 & 24.65       & 16.41       & 15.45             & 11.72           & 42.44 & 56.01  & 50.66  \\
                          & $L_{rce}$  & 17.69 & 25.33       & 17.33       & 16.36             & 12.56           & 43.11 & 68.57  & 56.05 \\
\hline
\multirow{6}{*}{COCO}     & $L_{fff}$  & 48.28 & 22.49       & 15.71       & 14.22             & 10.26           & 28.03 & 38.31  & 26.00 \\
                          & $L_{cos}$  & 21.71 & 25.92       & 17.38       & 16.87             & 12.93           & 45.02 & 71.47  & 62.72 \\
                          & $L_{cw}$   & 25.12 & 17.23       & 13.92       & 12.86             & 9.82            & 33.86 & 41.77  & 33.74 \\
                          & $L_{nce}$  & 20.26 & 23.91       & 15.98       & 15.08             & 11.69           & 42.00 & 51.66  & 45.35 \\
                          & $L_{cell}$ & 16.82 & 24.40       & 17.19       & 15.80             & 12.44           & 40.16 & 62.57  & 49.09 \\
                          & $L_{rce}$  & 16.74 & 24.56       & 17.13       & 15.72             & 12.41           & 42.24 & 65.40  & 51.89 \\
\hline         
\end{tabular}
\caption{Noise-based Framework}
\end{subtable}
\begin{subtable}[t]{\linewidth}
\centering
\begin{tabular}{c|c|cccccccc}
\hline
\multirow{2}{*}{Data} & \multirow{2}{*}{Loss} & Input Processing & \multicolumn{4}{c}{Adversarial Training}                        & \multicolumn{2}{c}{Stylized} & \multirow{2}{*}{Augmix} \\
                               &                       & NRP              & $L_{2}=0.1$ & $L_{2}=0.5$ & $L_{\infty} =0.5$ & $L_{\infty} =1$ & SIN          & SIN-IN        &                       \\
                               \hline
                               \multirow{6}{*}{ImageNet} & $L_{fff}$  &   40.66     & 26.75       & 12.70       & 12.07             & 8.46            & 52.42 & 68.84  & 60.19       \\
                          & $L_{cos}$  & 21.21 & 38.32       & 21.23       & 19.62             & 13.76           & 54.71 & 70.97  & 66.53  \\
                          & $L_{cw}$   & 22.55 & 47.97       & 20.33       & 18.26             & 12.35           & 58.24 & 73.54  & 64.74  \\
                          & $L_{nce}$  & 15.02 & 29.57       & 16.96       & 16.14             & 12.47           & 50.64 & 60.84  & 60.64  \\
                          & $L_{cell}$ & 23.86 & 42.14       & 18.61       & 17.98             & 13.12           & 57.58 & 77.35  & 67.65  \\
                          & $L_{rce}$  & 23.75 & 25.84       & 16.35       & 14.77             & 11.28           & 45.39 & 82.09  & 62.40 \\ \hline
\multirow{6}{*}{COCO}     & $L_{fff}$  &   48.24     & 25.15       & 12.44       & 11.35             & 8.14            & 41.11 & 73.91  & 54.15     \\
                          & $L_{cos}$  & 20.06 & 38.48       & 18.49       & 17.65             & 12.78           & 58.94 & 69.83  & 62.88 \\
                          & $L_{cw}$   & 21.05 & 46.57       & 21.27       & 18.58             & 12.65           & 56.56 & 64.22  & 58.89 \\
                          & $L_{nce}$  & 15.83 & 28.30       & 16.48       & 15.59             & 12.53           & 41.56 & 50.92  & 51.04 \\
                          & $L_{cell}$ & 25.67 & 43.43       & 19.15       & 18.03             & 13.16           & 57.81 & 77.95  & 66.24 \\
                          & $L_{rce}$  & 16.62 & 24.64       & 15.69       & 15.01             & 11.25           & 47.14 & 74.51  & 44.02 \\
\hline
\end{tabular}
\caption{Generator-based Framework}
\end{subtable}
\end{table*}

\subsection{Results in Targeted Attacks}
In this section, we compare different targeted attack approaches under the challenging 8-Targets setting. The 8 target classes are selected from the work of Inkawhich et al.~\cite{inkawhich2020transferable}. We evaluate their performance in both white-box and black-box settings using two metrics: the targeted fooling ratio (tFR) and the non-targeted fooling ratio (ntFR). The average values of these two metrics across the 8 targets are reported in Table~\ref{tab:target_white} (for white-box scenarios) and Table~\ref{tab:target_black} (for black-box scenarios).

\subsubsection{White-box attacks}
From Table~\ref{tab:target_white}, we can draw the following conclusions regarding the white-box targeted attack:
\textbf{1)} There is no significant difference in the targeted fooling ratio (tFR) when training with the proxy dataset (MS-COCO) compared to training with the original ImageNet dataset for generating targeted adversarial examples in both the noise-based and generator-based frameworks.
\textbf{2)} The noise-based framework achieves higher tFR than the generator-based algorithm across all three different loss functions, while the generator-based framework demonstrates a higher non-targeted fooling ratio (ntFR).
\textbf{3)} In the noise-based framework, the margin-based loss $L_{cw}$ achieves a higher tFR compared to the entropy-based losses $L_{ce}$ and $L_{rce}$. However, in the generator-based framework, $L_{ce}$ and $L_{rce}$ outperform $L_{cw}$ in terms of tFR. Notably, there are no significant differences in performance among these three loss functions.

\subsubsection{Black-box attacks} 
Based on the results presented in Table~\ref{tab:target_black}, we can draw similar conclusions regarding the black-box transferability of non-targeted attacks:
\textbf{1)} The targeted perturbation learned by the ResNet-50 can successfully attack more samples to the targeted class, compared to those generated by VGG-19. Specifically, the targeted perturbations from VGG-19 exhibit low transferability to other models.
\textbf{2)} The performance of using the ImageNet and COCO datasets is comparable, further indicating that the original training data is unnecessary for crafting Universal Adversarial Perturbations (UAPs).

However, there are some differences when compared to the non-targeted attacks. In non-targeted attacks, entropy-based losses show higher transferability compared to the margin-based loss $L_{cw}$. In contrast, $L_{cw}$ demonstrates higher transferability in targeted attacks. These results suggest that logits play an important role in learning targeted perturbations.
Moreover, the generator-based UAP trained on ResNet-50 exhibits significantly higher tFR and ntFR compared to the noise-based UAP. The generator-based UAP also achieves slightly better performance than the noise-based UAP on VGG-19.

\subsection{Results on unknown defense mechanisms}

In this section, we evaluate the robustness of the ResNet-50 trained by different defense strategies against six losses in non-targeted attacks (learned from the naturally trained ResNet-50 model). We choose four representative defense strategies from two mechanisms: input transformation and robust training. 

\textbf{Input Transformation.} Neural representation purifier (NRP)~\cite{naseer2020self} is a state-of-the-art defense method that projects perturbed images near the perceptual space of clean images through a purifier network (generative model)  before feeding them to the classifier. 

\textbf{Robust Training.} 
The fundamental principle of robust training is to retrain CNN models to resist various adversaries. We evaluate the performance of UAPs using three robust training methods: augmentation-based training, stylization-based training, and adversarial training.
The augmentation-based Augmix~\cite{hendrycks2019augmix} trains the CNN model by using the mixup version of the samples from several different data augmentations, which can make the model robust against natural corruptions. 
In stylized-based training, SIN~\cite{geirhos2018imagenet} re-trains the CNNs with the stylized ImageNet dataset, and SIN-IN~\cite{geirhos2018imagenet} re-trains models with a mixture of stylized ImageNet and the original ImageNet, which can improve understanding of CNN representations and shape bias.
Adversarial training~\cite{salman2020adversarially} can increase the robustness by leveraging the adversarial samples for training. 

We report the fooling rates of attacking the above defense mechanisms in Tab.~\ref{tab:nt-defense} and can observe the following phenomenons. 
\textbf{1)} By comparing these defense mechanisms, adversarial training can achieve lower fooling rates, which is superior to others. Besides, training with adversarial samples learned by the $L_\infty$ is more robust in resisting the UAPs.
\textbf{2)} In the generator-based and noise-based frameworks, attacking the stylization-based SIN and SIN-IN defenses results in higher fooling rates for the generator-based UAP compared to the noise-based UAP. However, there is no significant difference observed for other defense mechanisms.
\textbf{3)} Consistent with the findings in attacks, the UAPs learned from the original ImageNet and the proxy COCO have no significant difference from each other. These further suggest that the UAPs are more model-related and negligibly data-related.
\textbf{4)} Analyzing the different loss functions, we can find that the $L_{cos}$ is with the overall highest fooling rate in both the generator-based and noise-based frameworks. 
\begin{figure}[t]
    \centering
    \begin{subfigure}[b]{0.48\linewidth}
         \centering
         \includegraphics[width=\textwidth]{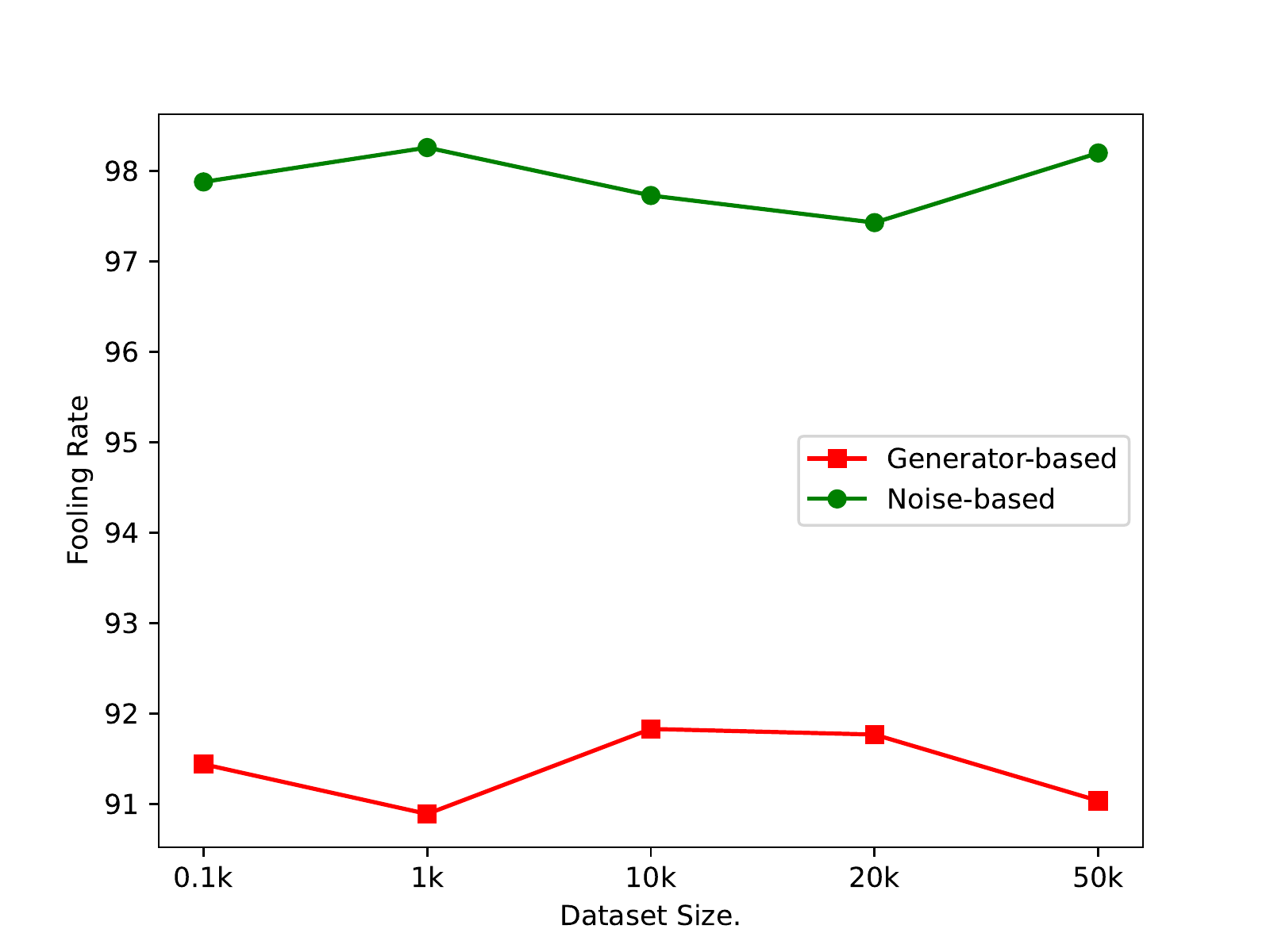}
         \caption{Non-target attack}
         \label{fig:size_non-target}
     \end{subfigure}
     \hfill
    \begin{subfigure}[b]{0.48\linewidth}
         \centering
         \includegraphics[width=\textwidth]{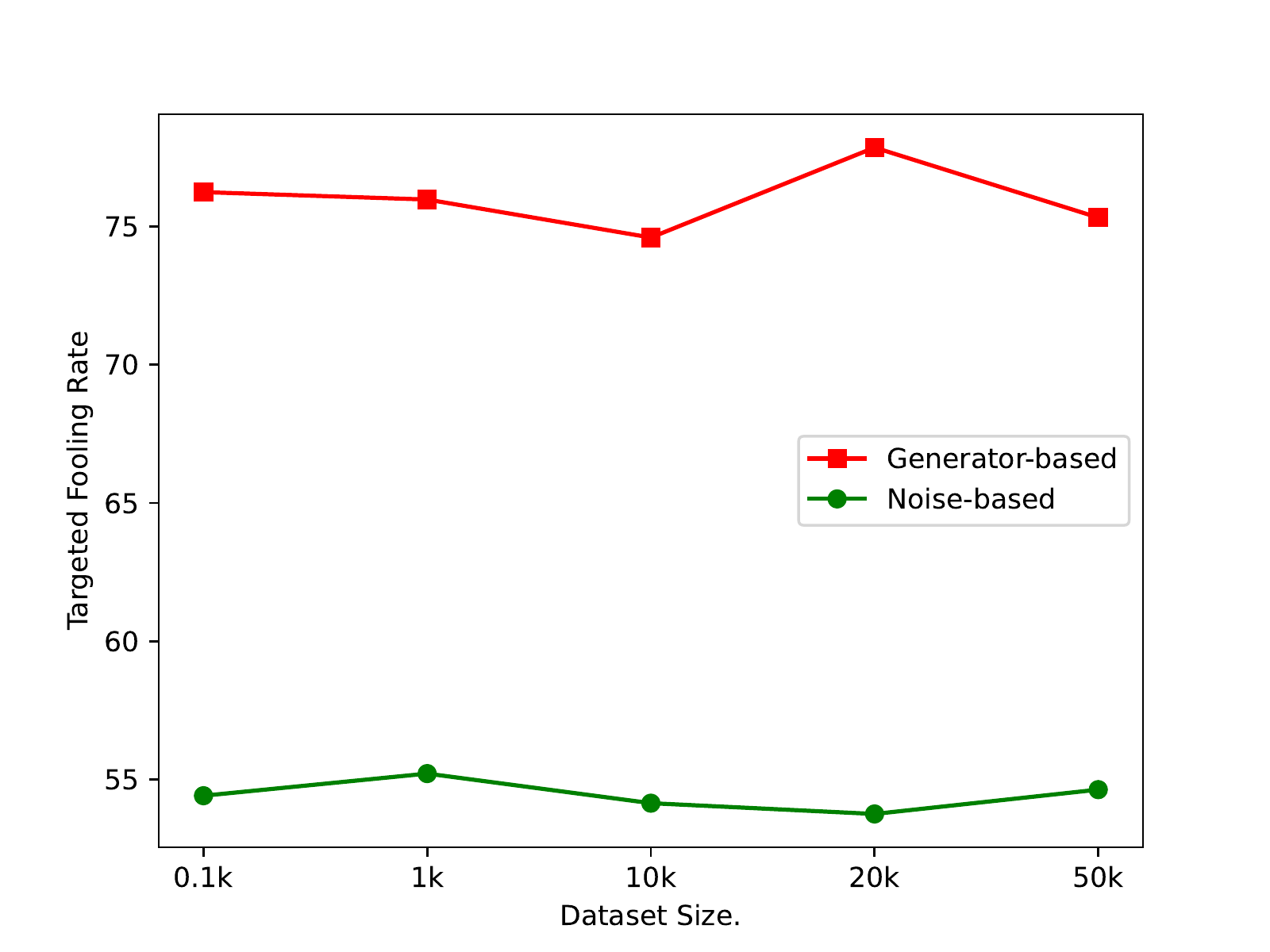}
         \caption{Target attack}
         \label{fig:size_target}
     \end{subfigure}
    % \subfigure[Targeted attacks]{
    % \includegraphics[width=0.3\linewidth]{figure/targted-attack.jpg}
    % \label{tab:model2}
    % }
    \caption{The influence of different amounts of training data. 
    %The universal attackers are crafted with a limited number of training samples (ImageNet data) for a ResNet-50 network. For the non-targeted attack, we adopt $L_{nce}$ loss to train the F-UAP and GAP. For the targeted attack, we use $L_{ce}$ loss to learn targeted perturbation of `grey owl’.
    }
    \label{fig:data-size}
\end{figure}

\begin{figure*}[t]
    \centering
    \includegraphics[width=0.8\linewidth]{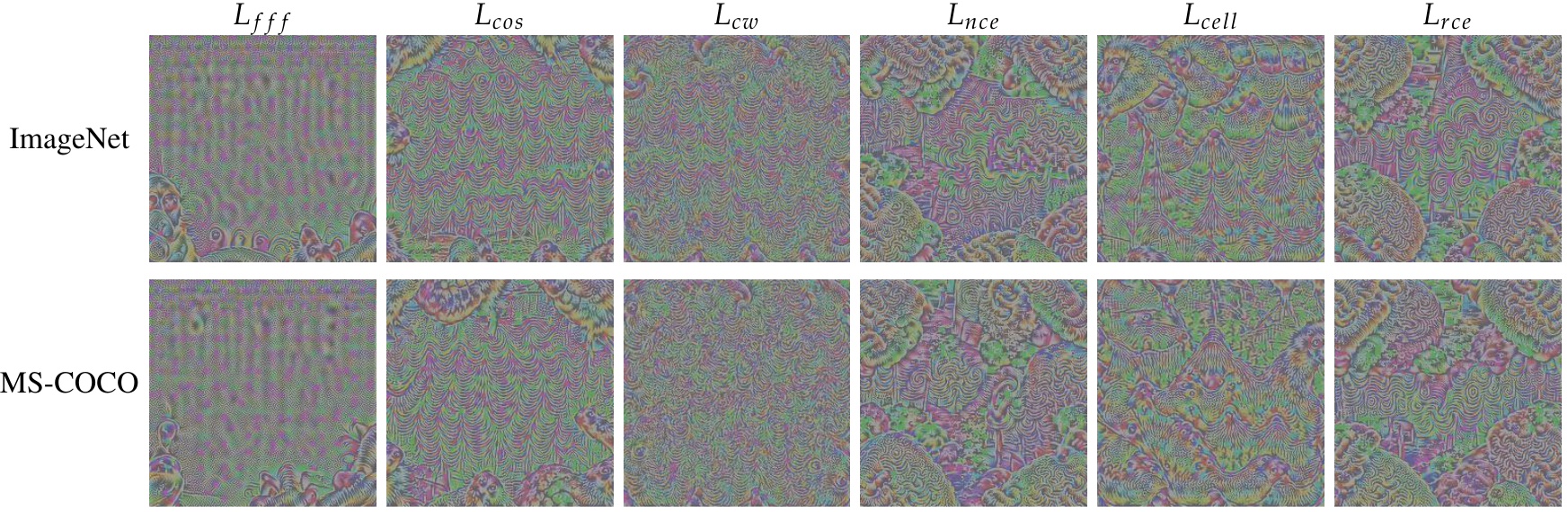}
    \caption{The non-targeted universal adversarial perturbations trained on different loss functions for the noise-based algorithm.}
    \label{fig:uap-non-target}
\end{figure*}

\begin{figure*}[!htbp]
    \centering
    \begin{subfigure}[b]{0.8\textwidth}
         \centering
         \includegraphics[width=\textwidth]{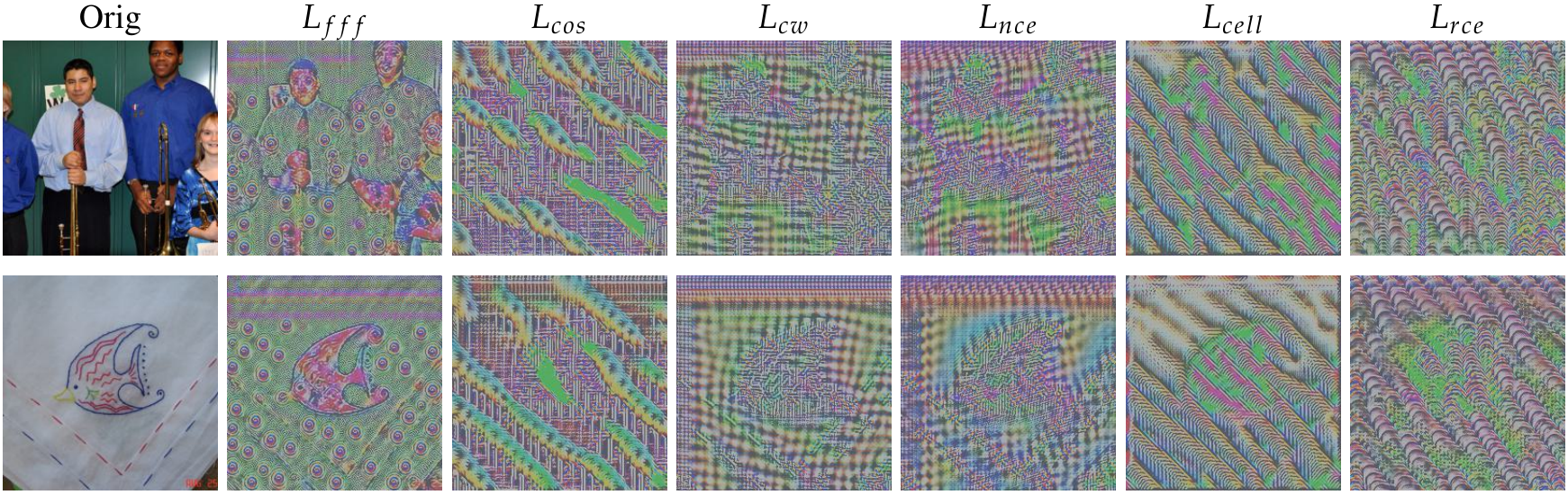}
         \caption{ImageNet}
     \end{subfigure}
     \hfill
    \begin{subfigure}[b]{0.8\textwidth}
         \centering
         \includegraphics[width=\textwidth]{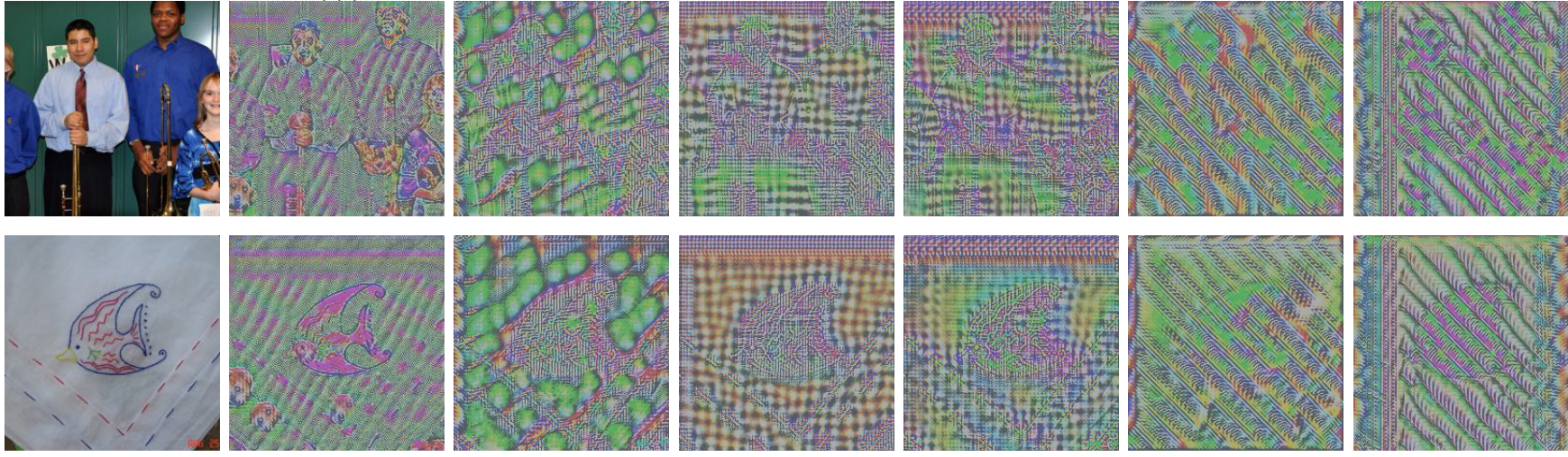}
         \caption{MS-COCO}
     \end{subfigure}
    \caption{The non-targeted adversarial perturbations trained on different loss functions for the generator-based algorithm.}
    \label{fig:gan-non-target}
\end{figure*}

\subsection{The influence of different amounts of training samples.}
In this part, we investigate the influence of the size of the training data. We select the ImageNet training set as our training data for two types of attack methods in order to fool the ResNet-50 model in the white-box setting.
For non-targeted attacks, we use the $L_{nce}$ loss to train both the generator-based UAP and the noise-based UAP. For targeted attacks, we employ the $L_{ce}$ loss to learn the targeted perturbation for the category "grey owl".
The results depicted in Figure~\ref{fig:data-size} demonstrate that the generator-based and noise-based frameworks are minimally affected by the varying amounts of training samples. Even with only $0.1k$ training samples, higher fooling rates can be achieved for both non-targeted and targeted attacks. These findings further verify that UAPs are more related to the model architecture and are agnostic to the input data.

% Tab.~\ref{tab:dataset-size} show that the GD-UAP, F-UAP, GAP and CD-TAP also performs well under limited access to data samples. Even with only $1,000$ training samples, a higher the fooling ratio can be achieved. For example, the fooling rates of the F-UAP and the CD-TAP are $95.96\%$ and  $99.04\%$ respectively. However, the attack performance of the UAP increases with more training data, which means that the UAP requires a large amount of data to train well on universal adversarial perturbations. For instance, when the training data size is increased to $20,000$, the fooling rate is $92.17\%$ and reaches saturation. 

\begin{figure*}[h]
    \centering
   \includegraphics[width=0.8\linewidth]{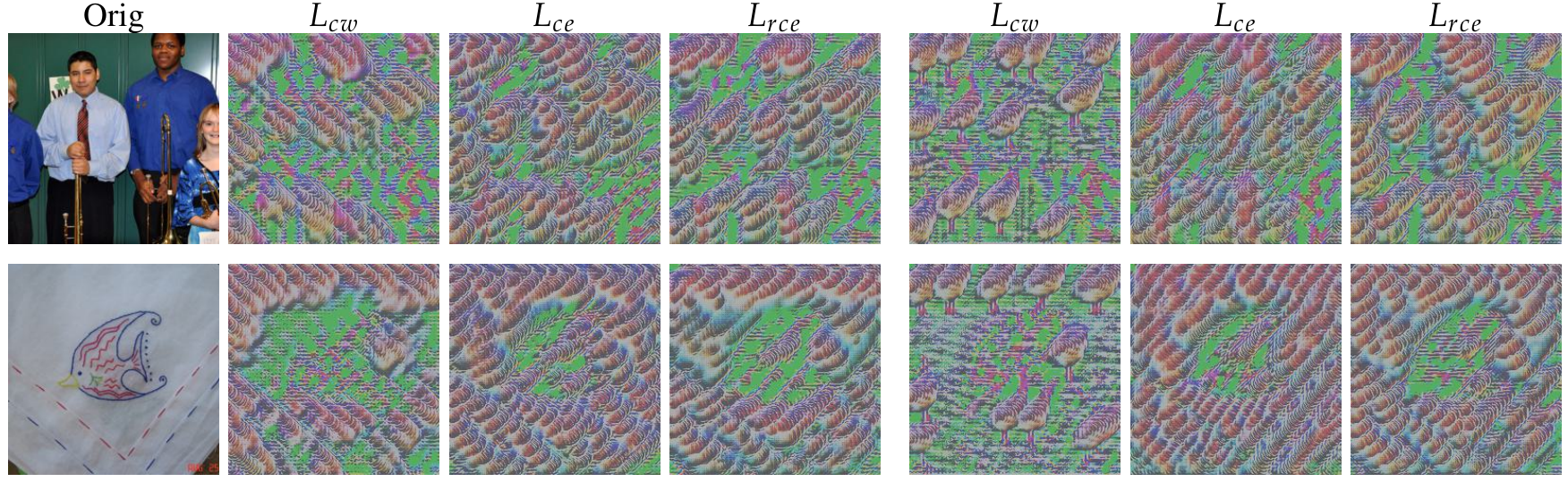}
    \caption{The targeted adversarial perturbations trained on different loss functions for the generator-based framework. The targeted class is `goose'.}
    \label{fig:gan-target}
\end{figure*}

\begin{figure}[h]
    \centering  
    \includegraphics[width=0.92\linewidth]{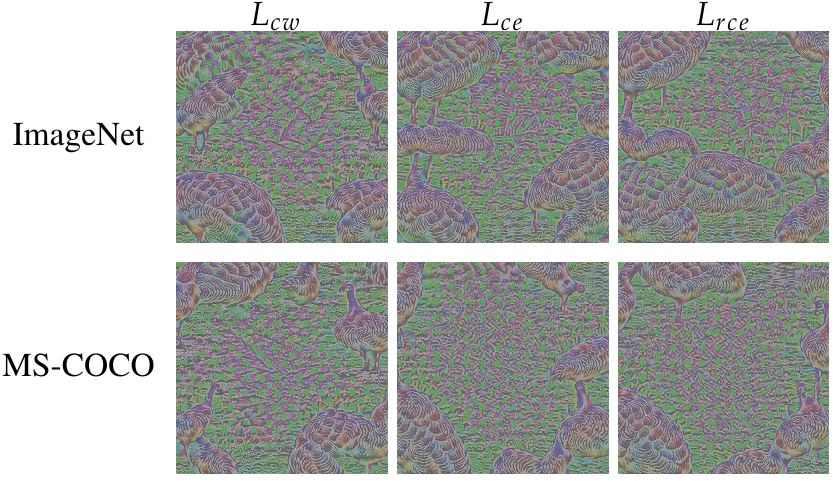}
    \caption{The targeted universal adversarial perturbations trained on different loss functions for the noise-based framework (Targeted class: `goose')}
    \label{fig:uap-target}
\end{figure}

% \begin{figure*}[t]
%     \centering
%     \begin{subfigure}[b]{0.5\linewidth}
%          \centering
%          \includegraphics[]{figure/gan-target-img.pdf}
%          \caption{ImageNet}
%      \end{subfigure}
%      \hfill
%     \begin{subfigure}[b]{0.4\linewidth}
%          \centering
%          \includegraphics[]{figure/gan-target-coco.pdf}
%          \caption{MS-COCO}
%      \end{subfigure}
%     \caption{The targeted adversarial perturbations trained on different loss functions for GAP. The targeted class is `goose'.}
%     \label{fig:gan-target}
% \end{figure*}

\section{Visualization and Discussion}
\label{sec:future}
In this section, we first visualize the generated UAP in both Non-target attacks and Target attacks. 
Then, we analyze the Pearson Correlation Coefficient in the target UAP from different models and loss functions.
Finally, we provide a brief discussion of future directions from the following two aspects, \textit{i.e.}, feature perspective, and geometric perspective.

\subsection{Visualization of the UAPs}

\subsubsection{Non-target attacks}
In Fig.~\ref{fig:uap-non-target}, we visualize the UAPs learned by the noise-based framework with different loss functions from the ImageNet and COCO datasets. First, we can find that the patterns of different losses vary from each other. But, the patterns of the same loss trained from the two datasets are very similar. On the other aspects, we also notice that the texture is uniform in the middle area of the UAPs, while the border contains more object-related texture. We argue this phenomenon is mainly due to the fact that foreground objects usually appear in the middle of the input image.

For the generator-based perturbations, as shown in Fig.~\ref{fig:gan-non-target}, we firstly can obtain similar findings about the losses and datasets as same as the noise-based UAP in Fig.~\ref{fig:uap-non-target}. Besides, we observe that the perturbation in the flat areas differs from the areas with more texture in the original input images. This means the generator-based method will add different perturbations in the foreground and background areas. Notably, the generator is universal and applicable to different input images.

\subsubsection{Target attacks}
The targeted perturbations are depicted in Figure~\ref{fig:uap-target} for the noise-based framework and in Figure~\ref{fig:gan-target} for the generator-based framework. From Figure~\ref{fig:uap-target}, we observe that the patterns are very similar across the three different loss functions. The high-frequency patterns associated with the target class are predominantly present in the border areas of the UAPs.
As for the generator-based UAP in Figure~\ref{fig:gan-target}, the high-frequency patterns tend to be added in the non-foreground-related areas with less texture. Considering the visualization of both non-targeted and targeted attacks, we can conclude that the perturbation is highly related to the textured areas of the foreground.

\subsection{Pearson Correlation Coefficient Analysis}
In this part, we analyze the correlations between the target UAP trained by different models and loss functions. Following~\cite{zhang2020understanding}, we compute the Pearson Correlation Coefficient (PCC) between the logits of different target UAPs by using the DenseNet-121. The target UAPs include the UAPs trained by $L_{rce}$ and $L_{rcw}$ for the ResNet-50 and the UAP trained by $L_{rce}$ for the VGG-19. 

\begin{figure}[t]
    \centering
    \begin{subfigure}[b]{0.49\textwidth}
         \centering
         \includegraphics[width=\textwidth]{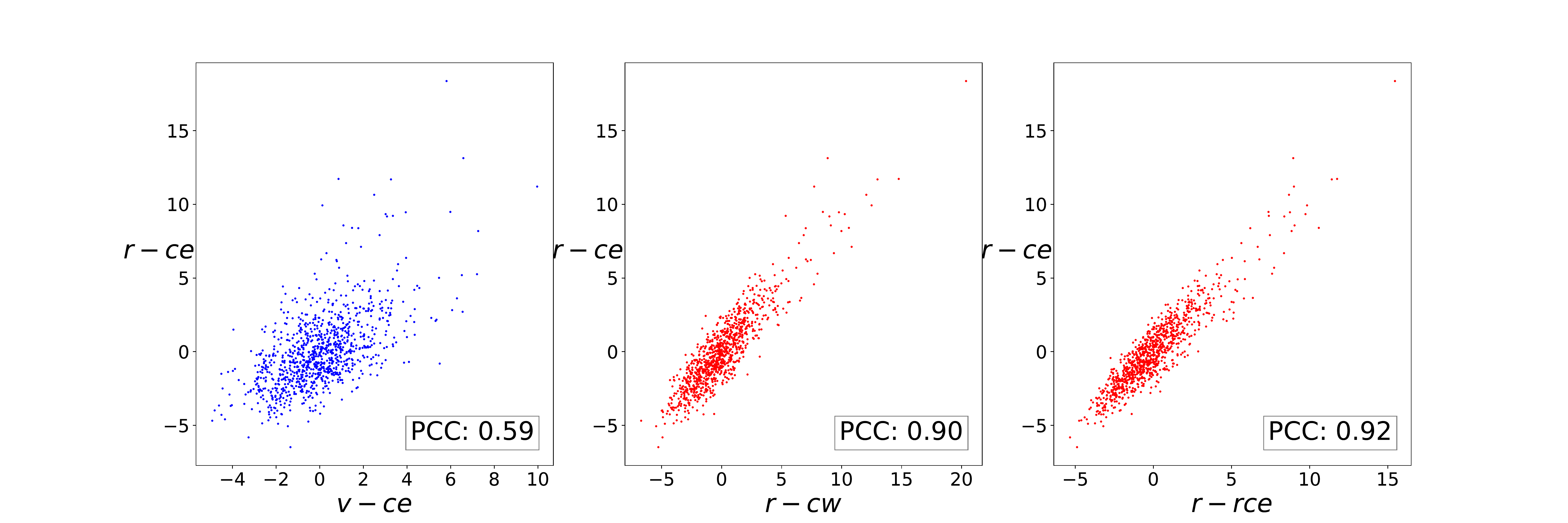}
         \caption{ImageNet}
     \end{subfigure}
     \hfill
    \begin{subfigure}[b]{0.49\textwidth}
         \centering
         \includegraphics[width=\textwidth]{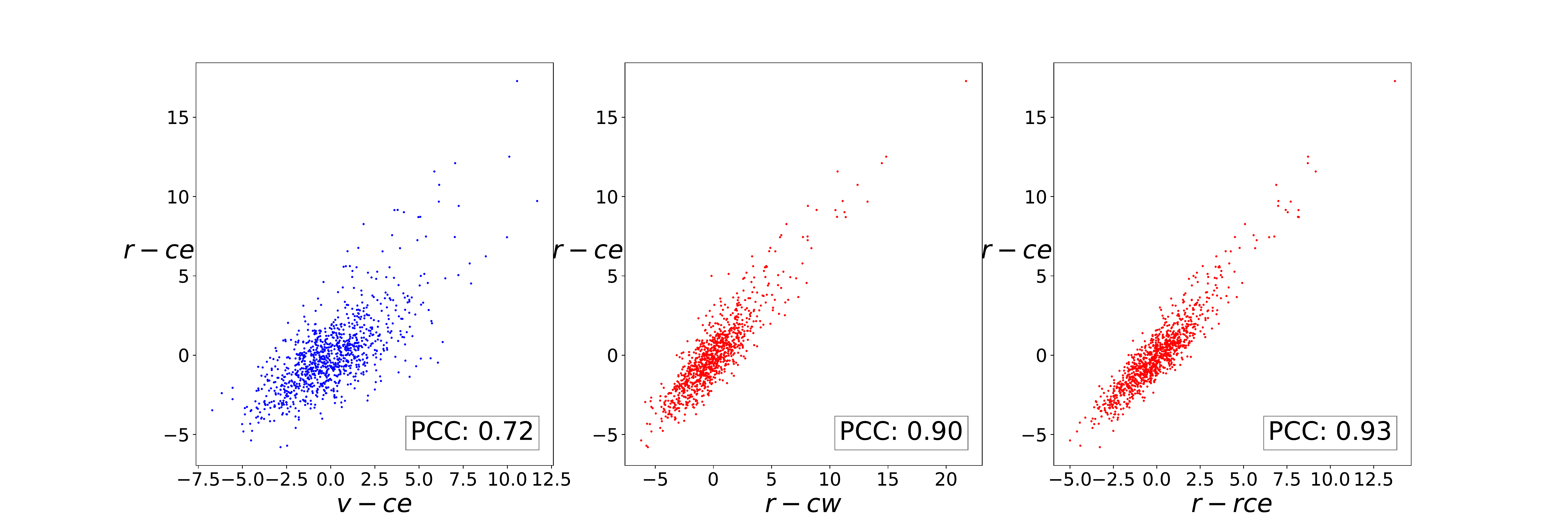}
         \caption{MS-COCO}
     \end{subfigure}
    \caption{The correlation analysis about targeted UAP from ImageNet data (a) and MS-COCO data (b). The targeted class is the `goose'. The `r-ce' and `v-ce' mean the UAPs are trained on ResNet-50 and VGG-19 with $L_{ce}$ loss functions, respectively. The `r-cw' and `r-rce' represent the UAPs are crafted with $L_{cw}$ and $L_{rce}$ loss, respectively.}
    \label{fig:pcc-ours}
\end{figure}

From Fig.~\ref{fig:pcc-ours}, we can have the following findings. 1) The PCCs with different loss functions for the same training model are quite high (\textit{e.g.}, $P(l(\delta_{r-ce}),  l(\delta_{r-cw}))=0.9$ and $P(l(\delta_{r-ce}),  l(\delta_{r-rcw}))=0.92$ for ImageNet), which means the targeted UAP is insensitive to different loss functions for crafting the same training model. 2) The PCCs with different training models are significantly lower than the PCCs with different loss functions, which means the UAPs are more relevant to the training model and the UAPs from different models are not similar. Besides, recently study~\cite{moosavi2017universal} suggests that the target UAP trained from a white-box model can capture the information of decision boundary against the model.

\subsection{Discussion}
\begin{itemize}
\item From a feature perspective, the existence of adversarial examples can be attributed to the presence of non-robust features~\cite{ilyas2019adversarial}, which poses a significant threat to real-world applications relying on deep models. Additionally, \cite{zhang2021universal} also indicates that the frequency is a key factor for the success of the universal adversarial attack, which can be attributed to the high sensitivity of deep models to high-frequency features. Hence, it is urgent to study defense strategies for creating robust models to alleviate adversarial attacks in the future. On the other aspect, we also notice that UAP is also related to the foreground area in the original image as shown in the visualization.
Therefore, we believe various defense methods~\cite{liu2018feature,xie2019feature} can be motivated by the high-frequency property and the consideration of object-related foreground areas.
    
\item 
From a geometric perspective, the UAPs are crafted from a subspace of low dimension, which captures the correlation between different parts of the decision boundary~\cite{moosavi2017universal} and the targeted UAPs are highly related to training models as shown in Fig.~\ref{fig:pcc-ours}. In view of this perspective, the UAPs can be applied for detecting model extraction attacks~\cite{tramer2016stealing,zhu2021hermes}. These attacks are to collect the outputs of the target model for training a piracy model, which means the piracy models are likely to inherit the decision boundary of the target model and the UAP subspace of piracy models may be similar to that of the target model. We believe model extraction detection-based UAPs~\cite{peng2022fingerprinting} might be an interesting future direction for providing deeper insight into defense.
\end{itemize}

\section{Conclusion}
\label{sec:conclusion}
In this paper, we first summarize the basic definition of UAPs and the corresponding evaluation metrics for non-targeted and targeted attacks. Next, we present the recent progress of UAPs in the image classification task from two aspects: noise-based methods and generator-based methods. Specifically, the generator-based methods can be divided into input-agnostic and input-related ways. To comprehensively compare the loss functions and the framework used for crafting UAPs, we conduct several experiments by leveraging the noise-based and the generator-based framework for training under the same training protocol with various loss functions. After analyzing quantitative results, several important observations can be found. Finally, we conclude through visualization and present two future directions from the feature perspective and geometric perspective of the existence of UAPs. 

\section*{Acknowledgements}
This work is supported by the National Natural Science Foundation of China (No. 62076210, 62276221); the Natural Science Foundation of Fujian Province of China (No. 2022J01002).
\bibliographystyle{elsarticle-num}
\bibliography{ref}

\begin{thebibliography}{10}
\expandafter\ifx\csname url\endcsname\relax
  \def\url#1{\texttt{#1}}\fi
\expandafter\ifx\csname urlprefix\endcsname\relax\def\urlprefix{URL }\fi
\expandafter\ifx\csname href\endcsname\relax
  \def\href#1#2{#2} \def\path#1{#1}\fi

\bibitem{szegedy2015going}
C.~Szegedy, W.~Liu, Y.~Jia, P.~Sermanet, S.~Reed, D.~Anguelov, D.~Erhan,
  V.~Vanhoucke, A.~Rabinovich, Going deeper with convolutions, in: IEEE
  Conference on Computer Vision and Pattern Recognition, 2015, pp. 1--9.

\bibitem{huang2017densely}
G.~Huang, Z.~Liu, L.~Van Der~Maaten, K.~Q. Weinberger, Densely connected
  convolutional networks, in: IEEE Conference on Computer Vision and Pattern
  Recognition, 2017, pp. 4700--4708.

\bibitem{ren2015faster}
S.~Ren, K.~He, R.~Girshick, J.~Sun, Faster r-cnn: Towards real-time object
  detection with region proposal networks, Advances in neural information
  processing systems 28 (2015) 91--99.

\bibitem{redmon2018yolov3}
J.~Redmon, A.~Farhadi, Yolov3: An incremental improvement, arXiv preprint
  arXiv:1804.02767.

\bibitem{chen2017deeplab}
L.-C. Chen, G.~Papandreou, I.~Kokkinos, K.~Murphy, A.~L. Yuille, Deeplab:
  Semantic image segmentation with deep convolutional nets, atrous convolution,
  and fully connected crfs, IEEE Transactions on Pattern Analysis and Machine
  Intelligence 40~(4) (2017) 834--848.

\bibitem{long2015fully}
J.~Long, E.~Shelhamer, T.~Darrell, Fully convolutional networks for semantic
  segmentation, in: IEEE Conference on Computer Vision and Pattern Recognition,
  2015, pp. 3431--3440.

\bibitem{szegedy2013intriguing}
C.~Szegedy, W.~Zaremba, I.~Sutskever, J.~Bruna, D.~Erhan, I.~Goodfellow,
  R.~Fergus, Intriguing properties of neural networks, arXiv preprint
  arXiv:1312.6199.

\bibitem{goodfellow2014explaining}
I.~J. Goodfellow, J.~Shlens, C.~Szegedy, Explaining and harnessing adversarial
  examples, arXiv preprint arXiv:1412.6572.

\bibitem{wang2020towards}
Q.~Wang, B.~Zheng, Q.~Li, C.~Shen, Z.~Ba, Towards query-efficient adversarial
  attacks against automatic speech recognition systems, IEEE Transactions on
  Information Forensics and Security 16 (2020) 896--908.

\bibitem{wu2020audio}
J.~Wu, B.~Chen, W.~Luo, Y.~Fang, Audio steganography based on iterative
  adversarial attacks against convolutional neural networks, IEEE transactions
  on information forensics and security 15 (2020) 2282--2294.

\bibitem{zhong2020towards}
Y.~Zhong, W.~Deng, Towards transferable adversarial attack against deep face
  recognition, IEEE Transactions on Information Forensics and Security 16
  (2020) 1452--1466.

\bibitem{dong2019efficient}
Y.~Dong, H.~Su, B.~Wu, Z.~Li, W.~Liu, T.~Zhang, J.~Zhu, Efficient
  decision-based black-box adversarial attacks on face recognition, in:
  Proceedings of the IEEE/CVF Conference on Computer Vision and Pattern
  Recognition, 2019, pp. 7714--7722.

\bibitem{wang2022transferable}
Z.~Wang, Y.~Zheng, H.~Zhu, C.~Yang, T.~Chen, Transferable adversarial examples
  can efficiently fool topic models, Computers \& Security 118 (2022) 102749.

\bibitem{peng2021ensemblefool}
W.~Peng, R.~Liu, R.~Wang, T.~Cheng, Z.~Wu, L.~Cai, W.~Zhou, Ensemblefool: A
  method to generate adversarial examples based on model fusion strategy,
  Computers \& Security 107 (2021) 102317.

\bibitem{xie2019improving}
C.~Xie, Z.~Zhang, Y.~Zhou, S.~Bai, J.~Wang, Z.~Ren, A.~L. Yuille, Improving
  transferability of adversarial examples with input diversity, in: Proceedings
  of the IEEE/CVF Conference on Computer Vision and Pattern Recognition, 2019,
  pp. 2730--2739.

\bibitem{dong2019evading}
Y.~Dong, T.~Pang, H.~Su, J.~Zhu, Evading defenses to transferable adversarial
  examples by translation-invariant attacks, in: Proceedings of the IEEE/CVF
  Conference on Computer Vision and Pattern Recognition, 2019, pp. 4312--4321.

\bibitem{liu2016delving}
Y.~Liu, X.~Chen, C.~Liu, D.~Song, Delving into transferable adversarial
  examples and black-box attacks, arXiv preprint arXiv:1611.02770.

\bibitem{kurakin2016adversarial}
A.~Kurakin, I.~Goodfellow, S.~Bengio, Adversarial machine learning at scale,
  arXiv preprint arXiv:1611.01236.

\bibitem{dong2018boosting}
Y.~Dong, F.~Liao, T.~Pang, H.~Su, J.~Zhu, X.~Hu, J.~Li, Boosting adversarial
  attacks with momentum, in: Proceedings of the IEEE conference on computer
  vision and pattern recognition, 2018, pp. 9185--9193.

\bibitem{moosavi2017universal}
S.-M. Moosavi-Dezfooli, A.~Fawzi, O.~Fawzi, P.~Frossard, Universal adversarial
  perturbations, in: IEEE conference on computer vision and pattern
  recognition, 2017, pp. 1765--1773.

\bibitem{mopuri2017fast}
K.~R. Mopuri, U.~Garg, R.~V. Babu, Fast feature fool: A data independent
  approach to universal adversarial perturbations, arXiv preprint
  arXiv:1707.05572.

\bibitem{poursaeed2018generative}
O.~Poursaeed, I.~Katsman, B.~Gao, S.~Belongie, Generative adversarial
  perturbations, in: IEEE Conference on Computer Vision and Pattern
  Recognition, 2018, pp. 4422--4431.

\bibitem{naseer2019cross}
M.~M. Naseer, S.~H. Khan, M.~H. Khan, F.~Shahbaz~Khan, F.~Porikli, Cross-domain
  transferability of adversarial perturbations, Advances in Neural Information
  Processing Systems (2019) 12905--12915.

\bibitem{zhang2020understanding}
C.~Zhang, P.~Benz, T.~Imtiaz, I.~S. Kweon, Understanding adversarial examples
  from the mutual influence of images and perturbations, in: IEEE/CVF
  Conference on Computer Vision and Pattern Recognition, 2020, pp.
  14521--14530.

\bibitem{zhang2021data}
C.~Zhang, P.~Benz, A.~Karjauv, I.~S. Kweon, Data-free universal adversarial
  perturbation and black-box attack, in: IEEE/CVF International Conference on
  Computer Vision, 2021, pp. 7868--7877.

\bibitem{hayes2018learning}
J.~Hayes, G.~Danezis, Learning universal adversarial perturbations with
  generative models, in: IEEE Security and Privacy Workshops, 2018, pp. 43--49.

\bibitem{naseer2021generating}
M.~Naseer, S.~Khan, M.~Hayat, F.~S. Khan, F.~Porikli, On generating
  transferable targeted perturbations, in: IEEE/CVF International Conference on
  Computer Vision, 2021, pp. 7708--7717.

\bibitem{khrulkov2018art}
V.~Khrulkov, I.~Oseledets, Art of singular vectors and universal adversarial
  perturbations, in: IEEE Conference on Computer Vision and Pattern
  Recognition, 2018, pp. 8562--8570.

\bibitem{benz2020double}
P.~Benz, C.~Zhang, T.~Imtiaz, I.~S. Kweon, Double targeted universal
  adversarial perturbations, in: Asian Conference on Computer Vision, 2020.

\bibitem{zhang2020cd}
C.~Zhang, P.~Benz, T.~Imtiaz, I.-S. Kweon, Cd-uap: Class discriminative
  universal adversarial perturbation, in: The AAAI Conference on Artificial
  Intelligence, Vol.~34, 2020, pp. 6754--6761.

\bibitem{mopuri2018generalizable}
K.~R. Mopuri, A.~Ganeshan, R.~V. Babu, Generalizable data-free objective for
  crafting universal adversarial perturbations, IEEE Transactions on Pattern
  Analysis and Machine Intelligence 41~(10) (2018) 2452--2465.

\bibitem{liu2019universal}
H.~Liu, R.~Ji, J.~Li, B.~Zhang, Y.~Gao, Y.~Wu, F.~Huang, Universal adversarial
  perturbation via prior driven uncertainty approximation, in: IEEE/CVF
  International Conference on Computer Vision, 2019, pp. 2941--2949.

\bibitem{mopuri2018nag}
K.~R. Mopuri, U.~Ojha, U.~Garg, R.~V. Babu, Nag: Network for adversary
  generation, in: IEEE Conference on Computer Vision and Pattern Recognition,
  2018, pp. 742--751.

\bibitem{hashemi2020transferable}
A.~S. Hashemi, A.~B{\"a}r, S.~Mozaffari, T.~Fingscheidt, Transferable universal
  adversarial perturbations using generative models, arXiv preprint
  arXiv:2010.14919.

\bibitem{mao2020gap++}
X.~Mao, Y.~Chen, Y.~Li, Y.~He, H.~Xue, Gap++: Learning to generate
  target-conditioned adversarial examples, arXiv preprint arXiv:2006.05097.

\bibitem{mopuri2018ask}
K.~R. Mopuri, P.~K. Uppala, R.~V. Babu, Ask, acquire, and attack: Data-free uap
  generation using class impressions, in: European Conference on Computer
  Vision, 2018, pp. 19--34.

\bibitem{lin2014microsoft}
T.-Y. Lin, M.~Maire, S.~Belongie, J.~Hays, P.~Perona, D.~Ramanan,
  P.~Doll{\'a}r, C.~L. Zitnick, Microsoft coco: Common objects in context, in:
  ECCV, 2014.

\bibitem{simonyan2014very}
K.~Simonyan, A.~Zisserman, Very deep convolutional networks for large-scale
  image recognition, arXiv preprint arXiv:1409.1556.

\bibitem{he2016deep}
K.~He, X.~Zhang, S.~Ren, J.~Sun, Deep residual learning for image recognition,
  in: IEEE Conference on Computer Vision and Pattern Recognition, 2016, pp.
  770--778.

\bibitem{zhang2021survey}
C.~Zhang, P.~Benz, C.~Lin, A.~Karjauv, J.~Wu, I.~S. Kweon, A survey on
  universal adversarial attack, arXiv preprint arXiv:2103.01498.

\bibitem{chaubey2020universal}
A.~Chaubey, N.~Agrawal, K.~Barnwal, K.~K. Guliani, P.~Mehta, Universal
  adversarial perturbations: A survey, arXiv preprint arXiv:2005.08087.

\bibitem{gal2016dropout}
Y.~Gal, Z.~Ghahramani, Dropout as a bayesian approximation: Representing model
  uncertainty in deep learning, in: International Conference on Machine
  Learning, 2016, pp. 1050--1059.

\bibitem{deng2009imagenet}
J.~Deng, W.~Dong, R.~Socher, L.-J. Li, K.~Li, L.~Fei-Fei, Imagenet: A
  large-scale hierarchical image database, in: CVPR, 2009.

\bibitem{kingma2014adam}
D.~P. Kingma, J.~Ba, Adam: A method for stochastic optimization, arXiv preprint
  arXiv:1412.6980.

\bibitem{inkawhich2020transferable}
N.~Inkawhich, K.~J. Liang, L.~Carin, Y.~Chen, Transferable perturbations of
  deep feature distributions, arXiv preprint arXiv:2004.12519.

\bibitem{naseer2020self}
M.~Naseer, S.~Khan, M.~Hayat, F.~S. Khan, F.~Porikli, A self-supervised
  approach for adversarial robustness, in: IEEE/CVF Conference on Computer
  Vision and Pattern Recognition, 2020, pp. 262--271.

\bibitem{hendrycks2019augmix}
D.~Hendrycks, N.~Mu, E.~D. Cubuk, B.~Zoph, J.~Gilmer, B.~Lakshminarayanan,
  Augmix: A simple data processing method to improve robustness and
  uncertainty, arXiv preprint arXiv:1912.02781.

\bibitem{geirhos2018imagenet}
R.~Geirhos, P.~Rubisch, C.~Michaelis, M.~Bethge, F.~A. Wichmann, W.~Brendel,
  Imagenet-trained cnns are biased towards texture; increasing shape bias
  improves accuracy and robustness, arXiv preprint arXiv:1811.12231.

\bibitem{salman2020adversarially}
H.~Salman, A.~Ilyas, L.~Engstrom, A.~Kapoor, A.~Madry, Do adversarially robust
  imagenet models transfer better?, Advances in Neural Information Processing
  Systems 33 (2020) 3533--3545.

\bibitem{ilyas2019adversarial}
A.~Ilyas, S.~Santurkar, D.~Tsipras, L.~Engstrom, B.~Tran, A.~Madry, Adversarial
  examples are not bugs, they are features, arXiv preprint arXiv:1905.02175.

\bibitem{zhang2021universal}
C.~Zhang, P.~Benz, A.~Karjauv, I.~S. Kweon, Universal adversarial perturbations
  through the lens of deep steganography: Towards a fourier perspective, in:
  The AAAI Conference on Artificial Intelligence, Vol.~35, 2021, pp.
  3296--3304.

\bibitem{liu2018feature}
C.~Liu, J.~JaJa, Feature prioritization and regularization improve standard
  accuracy and adversarial robustness, arXiv preprint arXiv:1810.02424.

\bibitem{xie2019feature}
C.~Xie, Y.~Wu, L.~v.~d. Maaten, A.~L. Yuille, K.~He, Feature denoising for
  improving adversarial robustness, in: IEEE/CVF Conference on Computer Vision
  and Pattern Recognition, 2019, pp. 501--509.

\bibitem{tramer2016stealing}
F.~Tram{\`e}r, F.~Zhang, A.~Juels, M.~K. Reiter, T.~Ristenpart, Stealing
  machine learning models via prediction $\{$APIs$\}$, in: 25th USENIX Security
  Symposium, 2016, pp. 601--618.

\bibitem{zhu2021hermes}
Y.~Zhu, Y.~Cheng, H.~Zhou, Y.~Lu, Hermes attack: Steal $\{$DNN$\}$ models with
  lossless inference accuracy, in: 30th USENIX Security Symposium, 2021.

\bibitem{peng2022fingerprinting}
Z.~Peng, S.~Li, G.~Chen, C.~Zhang, H.~Zhu, M.~Xue, Fingerprinting deep neural
  networks globally via universal adversarial perturbations, in: IEEE/CVF
  Conference on Computer Vision and Pattern Recognition, 2022, pp.
  13430--13439.

\end{thebibliography}

\end{document}